\documentclass[12pt,journal,onecolumn,draftcls]{IEEEtran}
\ifCLASSINFOpdf
\else
\fi

\usepackage{cite}
\usepackage{amsmath,amssymb,amsfonts}
\usepackage{algorithmic}
\usepackage{graphicx}
\usepackage{textcomp}
\usepackage{xcolor}
\usepackage{url}
\usepackage{subcaption}
\usepackage[font=small,labelfont=bf]{caption}

\DeclareMathOperator*{\argmin}{arg\,min}
\usepackage[ruled]{algorithm2e}
\usepackage{diagbox}
\usepackage{multirow}
\usepackage{color}
\begin{document}
%
\title{Towards Semantic Communications: \\ Deep Learning-Based Image Semantic Coding}
%
%
%

\author{Danlan~Huang,~\IEEEmembership{Member,~IEEE,}
        Feifei~Gao,~\IEEEmembership{Fellow,~IEEE,}
        Xiaoming Tao,~\IEEEmembership{Member,~IEEE,}
        Qiyuan~Du,
        ~and~Jianhua~Lu,~\IEEEmembership{Fellow,~IEEE}
        
\IEEEauthorblockN{
}
\thanks{D. Huang, F. Gao, X. Tao, Q. Du and J. Lu are with Beijing National Research Center for Information Science and Technology, Tsinghua University, Beijing, China. X. Tao is the corresponding author. Email:\{huangdl, feifeigao, taoxm, lhh-dee\}@mail.tsinghua.edu.cn. }
}
\maketitle
\vspace{-15mm}
\begin{abstract}
Semantic communications has received growing interest since it can remarkably reduce the amount of data to be transmitted without missing critical information. Most existing works explore the semantic encoding and transmission for text and apply techniques in Natural Language Processing (NLP) to interpret the meaning of the text. In this paper, we conceive the semantic communications for image data that is much more richer in semantics and bandwidth sensitive. We propose an reinforcement learning based adaptive semantic coding (RL-ASC) approach that encodes images beyond pixel level. Firstly, we define the \textit{semantic concept} of image data that includes the category, spatial arrangement, and visual feature as the representation unit, and propose a convolutional semantic encoder to extract semantic concepts. Secondly, we propose the image reconstruction criterion that evolves from the traditional \textit{pixel similarity} to \textit{semantic similarity} and \textit{perceptual performance}.  Thirdly, we design a novel RL-based semantic bit allocation model, whose reward is the increase in \textit{rate-semantic-perceptual} performance after encoding a certain semantic concept with adaptive quantization level. Thus, the task-related information is preserved and reconstructed properly while less important data is discarded.  Finally, we propose the Generative Adversarial Nets (GANs) based semantic decoder that fuses both locally and globally features via an attention module. Experimental results demonstrate that the proposed RL-ASC is noise robust and could reconstruct visually pleasant and semantic consistent image, and saves times of bit cost compared to standard codecs and other deep learning-based image codecs.
\end{abstract}

\begin{IEEEkeywords}
Semantic communications, image semantic coding, Generative Adversarial Nets (GANs), reinforcement learning (RL),  rate-semantic-perceptual criterion
\end{IEEEkeywords}


%
\IEEEpeerreviewmaketitle

\section{Introduction}
%
%
%
%

\IEEEPARstart{D}ating back to 1949, Weaver \cite{1949The} defined three levels of communications: the bit level, the semantic level, and the effective level. The conventional communications falls into the bit level, where the receiver recovers the raw data by maximizing the accuracy of symbol transmission; The semantic communications interprets information at the semantic level and attempts to transmit the symbol that precisely conveys the desired meaning, instead of bit copy;  The effective level concerns how effectively would the received meaning affect the behavior of the receiver in the desired way. Recently, deep learning achieves a breakthrough in semantic analysis task such as NLP, image processing, and speech recognition, etc, and therefore it paves the way for building the semantic communication systems and enables intelligent communication for human-to-machine as well as machine-to-machine. 

Recent works mainly focus on semantic communications for text/speech modalities  \cite{XieTSP,Xie,Weng,DBLP:journals/corr/abs-2201-01389}, and therefore can be categorize as \textit{text semantic communications}(TSC). A deep learning based joint source-channel coding (JSCC) of text is proposed in \cite{8461983}, which achieves lower word error rate and preserves semantic information of sentences. Later, the authors of \cite{lu2021reinforcement} design an reinforcement learning (RL) based semantic communication system to deal with the non-differentiability of semantic metrics and interact with the surrounding noisy environment.



Actually, the vision modal is more informative than text modal, and the video/image traffic accounts for 75\% of IP traffic nowadays \cite{Cisco2017Cisco}. However, the huge amount of multimedia traffic encounters the transmission challenges such as delay and network congestion, which are difficult to be solved by the conventional communication techniques. Hence, it is more important to build the \textit{image semantic communications} (ISC) that can remarkably reduce the amount of data to be transmitted without sacrificing the semantic fidelity of the image.

The earlier attempt of semantic image communications \cite{Bourtsoulatze,Kurka,XuTCSVT} developed the JSCC method and achieved good performance in the challenging low signal-to-noise (SNR) and small bandwidth regimes.  The work \cite{Bourtsoulatze} maps the image pixel values directly to the complex-valued channel input symbols and learns noise resilient coded representations, and therefore outperforms separation-based digital communication at all SNR. Later, Kurka et.al \cite{Kurka} proposed an multiple-description JSCC scheme for bandwidth-agile image transmission. The attention DL based JSCC for image transmission \cite{XuTCSVT} successfully operates with different SNR levels during transmission.

However, the aforementioned works lack the interpretation ability of the image content. Actually, the main issue of the ISC is to discard the goal of precise reconstruction and pursue semantic fidelity of the reconstructed image even in aggressive compression ratio. The bit level image communications such as the standard codecs \cite{JPEG,JPEG2000,BPG} and deep learning based codecs \cite{20181304961211, 20191106642196,2018Nips,2015Variable,2018Conditional} aim to solve the rate-distortion optimization (RDO) problem. They exactly recover the transmitted image data at the receiver side by processing the image in pixel level, and therefore gradually approach the compression limits of Shannon's information theory. The highly compressed image essentially deviate from human perception and suffers from degradation such as blocking, ringing, blurry, and checkerboard artifacts \cite{Agustsson,johnson2016perceptual} that cause poor performance in semantic analysis tasks such as classification, detection, and segmentation \cite{Grm,7498955}.  
 Therefore, blindly minimizing pixel-wise distortion may bring unnecessary bit overhead. The work \cite{LI202213} states that the conventional distortion metric Mean Squared
Error (MSE) in compression cannot fulfill the requirements of desirable intelligent task performance, and it is essential to exploit machine-centric evaluation metrics for high inference accuracy.


Some preliminary studies have tried to leverage the semantic similarity as the reconstruction criterion, which tolerates certain pixel-level errors and evaluates the usefulness of the reconstructed image in the sense that it better serves for the downstream semantic analysis task \cite{liu2019classification}.  The semantic similarity metric is successfully applied in low bitrate facial image compression \cite{torfason2018towards, chen2019learning}, where aggressive compression ratio is realized by filtering out the task irrelevant information. However, their supported analysis task is limited to face recognition and cannot be generalized to other applications. A task-driven semantic coding with the traditional hybrid coding framework integrates the semantic fidelity metric into the optimization process, and implements the semantic bit allocation based on reinforcement learning \cite{9472999}.  However, its traditional HEVC framework still process the image at pixel level and lacks the interpretation ability.
A detection-driven image compression with semantically structured bit-stream is proposed in \cite{2019Beyond}, where each part of the bitstream represents a specific object. Later on, the work \cite{9281078} generalizes the  semantically structured image coding framework to multiple intelligent tasks. However,  it follows the common practice in the traditional hybrid compression framework and utilize predictive coding to reconstruct the image in pixel-level. Torfason et al. \cite{torfason2018towards} verified that semantic analysis such as classification and segmentation, can be
performed on the bitstream directly without image decoding. However,  it cannot produce high quality natural image reconstruction, and the bitstream still describes the entire image without semantic structure.

Moreover, all of the above methods cannot meet the semantic and perception \cite{ICLR} performance requirements jointly in low bit rate.  Actually, different regions of the image vary in semantic importance and should be encoded adaptively with appropriate bitrate. Reinforcement learning is a promising method to extract the task-related information and adaptively encode different regions by selecting the optimal discrete quantization coefficients. Moreover, the adversarial loss of GANs \cite{goodfellow2014generative,mentzer2020high} captures the distribution of natural images and coincide with human perception. However, it is not feasible to directly adopt the naive GANs for image semantic decoding, since the generated image tend to deviate significantly from the input image in fine-grained features \cite{Santurkar,Agustsson,Akbari}.


In this paper, we propose an RL-based adaptive semantic coding (RL-ASC) approach to jointly address the semantic similarity and perceptual performance issues for ISC. The bandwidth burden could be remarkably alleviated due to the relaxing of precision requirements.  Our contribution are summarized as follows:
\begin{itemize}
    \item We propose a novel representation unit named \textit{semantic concept} that contains the category, feature and spatial relation of each object. A convolutional semantic encoder is proposed to extract semantic concepts at the transmitter and transform them into bit streams.
    \item We propose a task-driven image semantic coding framework that adopts new reconstruction criterion \textit{rate-semantic-perceptual} loss. We present an RL-based semantic bit allocation model to assign the optimal discrete quantization coefficients and achieve adaptive coding. 
    \item We propose an attention-based generative semantic decoder at the receiver to reconstruct the image with high perceptual quality in an aggressive compression ratio. Particularly, both global and local generators are learned to capture the global and category-wise features.
\end{itemize}

The rest of this work is organized as follows. In Section \uppercase\expandafter{\romannumeral2}, we propose the RL-ASC method to optimize the rate-semantic-perceptual criterion. In Section \uppercase\expandafter{\romannumeral3}, the soft quantizer, the network architecture of attention-based generative semantic decoder as well as the training algorithm are proposed.  The experiment details and the performance are shown in Section \uppercase\expandafter{\romannumeral4}, and the conclusions are made in Section \uppercase\expandafter{\romannumeral5}.

\section{The RL-based Adaptive Semantic Coding}
In this section, we propose the semantic concept as the representation unit and adopt semantic fidelity and perceptual quality as the optimization criterion. Next, we design an RL-based semantic bit allocation model to highlight and encode the task-related semantic concepts adaptively. 

\begin{figure*}[t]
\centering
\includegraphics[width=1.0\linewidth]{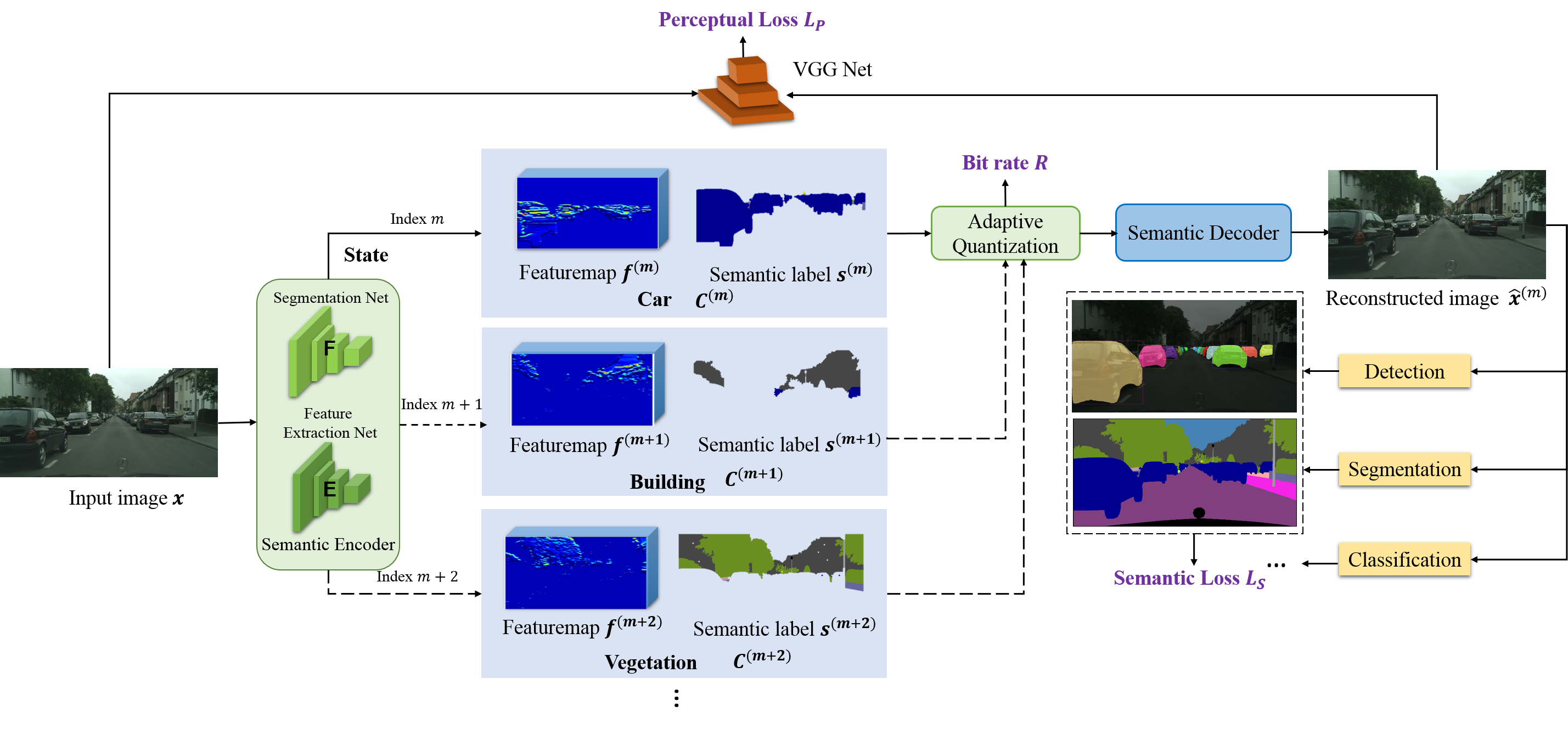}
\caption{System model of the proposed RL-based adaptive semantic coding method.}
\label{inference}
\end{figure*}

The system model of the proposed RL-ASC is illustrated in Fig. \ref{inference}, which mainly includes three modules: the semantic encoder, the agent $\pi$ that conducts adaptive quantization, and the semantic decoder. In the inference stage, the three modules are pretrained and fixed to accommodating to a certain downstream task. The semantic encoder extracts the semantic information of each class of the given input image. Then, each class of objects is assigned with an optimal quantization level learned by the RL agent $\pi$ to assure aggressive bit rate $R$. At the receiver, the semantic decoder reconstructs all the semantic information in parallel that performs well in both semantic loss $L_s$ and perceptual loss $L_p$.

\subsection{Representation Unit: From Pixel to Semantic Concept}

Existing image communications takes pixel as the representation unit and forces the decoded image to appear exactly the same as the transmitted one. On the contrary, the ISC interprets the underlying semantics of the image and thus is tolerant to certain pixel errors. Here we consider the urban street scene scenario and adopt the \textit{Cityscapes} dataset, where the quantity and classes of the semantic concepts are predefined. The dataset contains $M=30$ number of classes grouped into the following categories: flat surfaces, humans, vehicles, constructions, objects, nature, sky, and void. If the image details cannot fit into any classes, then it belongs to the void category.

Denote the input image as $\pmb{x}\in \mathbb{R}^{3\times H \times W}$, where $``3"$ denotes the RGB channels and $H$, $W$ are the dimensions along the height and width directions. We design the semantic encoder at the transmitter to extract semantic information for individual categories. The input image $\pmb{x}$ is fed into the segmentation network $\mathcal{F}$ to obtain semantic label map: $\pmb{s}=\mathcal{F}(\pmb{x})$, where $\pmb{s}\in \mathbb{R}^{W\times H}$ assigns a class label to each pixel in $\pmb{x}$ and thus reveals the spatial arrangement of the total $M$ objects. The entry of $\pmb{s}$ at pixel coordinate $(i,j)$ is an integer $\pmb{s}(i,j)\in \{1,2,...,M\}$ that represents the class label. We denote $\pmb{s}^{(m)}$ where $m=1,2,...,M$ is the label ID, as the \textit{semantic mask} for the $m$th semantic concept $\textbf{C}^{(m)}$, the entry of which is expressed as:
\begin{equation}
    \pmb{s}^{(m)}(i,j)= \left\{\begin{array}{ll}
    & 1, \text{ if $\pmb{s}(i,j)=m$,}\\
    & 0, \text{ otherwise}.
    \end{array}
    \right.
\end{equation}
To meet the transmission bit rate requirements, the semantic information should be down-scaled and then transform into bit stream. The down-scaled semantic mask is denoted as $\pmb{s}_d^{(m)}\in \mathbb{R}^{ w\times h}$, where $w\leq W, h \leq H$.

Meanwhile, $\pmb{x}$ is fed into the feature extraction network $\mathcal{E}$ that convolutionally processes $\pmb{x}$ into a feature map of size $N \times w \times h $. Then, this feature map is projected down to $n$ channels ($n<N$) to downscale the dimension and discard perceptually redundant features, in order to meet the bit rate requirements, resulting in a feature map $\pmb{f} \in \mathbb{R}^{n \times w\times h}$.  Note that each convolutional layer is followed by instance normalization and LeakyReLU activation.  The category-specific feature maps $\pmb{f}^{(m)}\in \mathbb{R}^{n \times w\times h}$ for $\textbf{C}^{(m)}$ is obtained by decomposing $\pmb{f}$: 
\begin{equation}
\pmb{f}^{(m)} = \pmb{f}\odot \pmb{s}_d^{(m)},
\end{equation}
where $\odot$ is the element-wise multiplication.

The filtered category-specific feature maps $\pmb{f}^{(m)}$ should be discriminative enough.  We design a novel classification module that adopts the feature classification loss $L_C$ to assure the distinctiveness. The architecture of the classification module is as follows. Firstly, all the $\pmb{f}^{(m)}$'s are fed into a max-pooling layer to yield $M$ pooled feature maps with dimension of $n\times 1\times 1$. Next, an fully connected (FC) layer with parameters shared across different $\pmb{f}^{(m)}$'s is adopted, and the outputs are $M$ logits with dimension $M\times 1$. The subsequent softmax function returns the predicted classification probability $\hat{p}^{(m)}$ for each category-specific feature maps $\pmb{f}^{(m)}$,  while the one-hot ground truth label is denoted as $p^{(m)}$. The feature classification loss $L_C$ can then be written as the cross entropy loss
\begin{equation}
    L_C = -\sum_{m=1}^M p^{(m)}\log\hat{p}^{(m)}. 
\end{equation}


In such a way, we imitate the image understanding process of humans and divide $\pmb{x}$ into $M$ individual semantic concepts that each can be denoted as a set $\textbf{C}^{(m)}=\{\pmb{f}^{(m)} \in\mathbb{R}^{n\times w\times h},\pmb{s}_d^{(m)}\in\mathbb{R}^{w\times h}\}$. Therefore,  $\textbf{C}^{(m)}$ can be deemed as novel representation unit that is far more efficient than pixel unit.

\subsection{Reconstruction Metric: From Pixel Loss to Semantic-Perceptual Loss}
Conventional image communications system measures the pixel loss in terms of PSNR and the coding process is optimized by the rate-distortion theory. However, the main purpose of ISC is to transmit the underlying meaning of the image other than pixel copy. The ISC should extract, encode, and reconstruct each $\textbf{C}^{(m)}$ with low bitrate in the sense that critical concepts are recovered with high semantic fidelity and perceptual quality. We propose a novel rate-semantic-perceptual criterion to optimize the semantic coding process.

It is already known that the intelligent analysis tasks such as object detection, semantic segmentation, pose estimation, and action recognition, etc., are able to extract semantic information accurately, and the obtained semantic information could also be used in the image coding process.  Denote the network of the intelligent task as $\mathcal{H}$, and then the prediction result of the input image $\pmb{x}$ and reconstructed image $\hat{\pmb{x}}$ can be written as  $\mathcal{H}(\pmb{x})$ and $\mathcal{H}(\hat{\pmb{x}})$. We propose a novel semantic loss $L_S$ metric defined as the degradation of the precision performance of the intelligent task $\mathcal{H}$ on the reconstructed image $\hat{\pmb{x}}$ compared to the original image $\pmb{x}$.  The $L_S$ can even be generalized to arbitrary user-defined semantic tasks, which can be  written as:
\begin{equation}
    L_S = f_{degrade}(\mathcal{H}(\pmb{x}), \mathcal{H}(\hat{\pmb{x}})),
\end{equation}
where $f_{degrade}$ is the accuracy degradation function of the predicted result given the ground truth. In such a way, we successfully convert abstract semantic information to a measurable form. 

The PSPNet \cite{Zhao} can be leveraged to implement the semantic segmentation task $\mathcal{H}$. while the bounding set of the predicted object of class $m$ on $\hat{\pmb{x}}$ is $\mathcal{H}(\hat{\pmb{x}})=\pmb{B}_d^m$, while the ground truth bounding set on $\pmb{x}$ is denoted as $\mathcal{H}(\pmb{x})=\pmb{B}_g^m$.  The performance of semantic segmentation can be evaluated by IoU that measures the 
consistency of $\pmb{B}_g^m$ and $\pmb{B}_d^m$:
\begin{equation}
    IoU(\mathcal{H}(\pmb{x}), \mathcal{H}(\hat{\pmb{x}})) = IoU(\pmb{B}_g^m, \pmb{B}_d^m) = overlap(\pmb{B}_g^m, \pmb{B}_d^m)/union(\pmb{B}_g^m, \pmb{B}_d^m),
\end{equation}
where IoU is a value between 0 and 1. Mean IoU (mIoU) is the average segmentation precision over the total $M$ classes. The semantic loss is calculated by mIoU and represents the extent of the non-interpretability of the reconstructed image $\hat{\pmb{x}}$:
\begin{equation}
    L_S = 1-\frac{1}{M}\sum_{m=1}^M IoU(\pmb{B}_g^m, \pmb{B}_d^m).
\end{equation}
The ideal $\pmb{\hat{x}}$ should preserve certain semantic information required by PSPNet, which results in $L_s=0$. 

The Mask R-CNN \cite{he2018mask} can be utilized to implement the object detection task $\mathcal{H}$. We adopt IoU loss \cite{Yu_2016} as the semantic loss:
\begin{equation}
    L_S=  -\ln[\frac{1}{M} \sum_{m=1}^M IoU(\pmb{B}_g^m, \pmb{B}_d^m)].
\end{equation}

For image classification task, the semantic loss $L_S$ is defined as the typical cross-entropy function. Let $\mathcal{H}$ be the pretrained classification network such as VGG and MobileNet \cite{2018MobileNetV2}. Let $\mathcal{H}(\pmb{x})=p_1,p_2,...,p_M$ be the one-hot vector of the ground truth label. Denote  $y$ as the ground truth label ID, and therefore we can obtain
\begin{equation}
    p_i = \left\{
    \begin{aligned}
    1,\quad & if (i=y)\\
    0,\quad & if (i \neq y)
    \end{aligned}
    \right.
\end{equation}
The predicted output is denoted as $\mathcal{\hat{\pmb{x}}}=\hat{p}_1, \hat{p}_2,...,\hat{p}_M$ corresponding to $M$ classes.The semantic loss is computed as
\begin{equation}
    L_{S}= -\frac{1}{M}\sum_{i=1}^M(p_i \log(\hat{p}_i)),
    \label{ce}
\end{equation}

For person re-identification, the performance can be evaluated by the cross-entropy loss \eqref{ce} utilizing the label smoothing technique. The true probability distribution is rewritten as
\begin{equation}
    p_i = \left\{
    \begin{aligned}
    1-\epsilon,\quad & if (i=y)\\
    \epsilon/(K-1),\quad & if (i \neq y)
    \end{aligned}
    \right.
\end{equation}
where $\epsilon$ is a small hyper parameter.

Besides, the visual performance such as the naturalness and clarity of $\pmb{\hat{x}}$ is another critical issue for the ISC.  The perceptual loss projects the images $\pmb{x}$ and $\hat{\pmb{x}}$ to a high dimensional feature space with a pretrained model and minimize the distance in the high-level space to measure the similarity of the two images. The perceptual loss is validated to predict human scores for degradation properly and could be expressed as:
 \begin{equation}\label{distortion2}
    L_P =  \|\phi(\pmb{x}) - \phi(\pmb{\hat{x}})\|_2^2,
\end{equation}
where $\phi(\cdot)$ is the pretrained VGG network that maps an input image to a high dimension feature space.

We then formulate the rate-semantic-perceptual criterion $L$ as the triple trade-off between semantic fidelity $L_S$, perceptual quality $L_P$, and bit rates $R$ of the semantic concepts:
\begin{equation}\label{triple}
    L = \lambda R + L_S + \eta L_P
\end{equation}
where $R$ is measured by the length of transmitted bit stream, and $\lambda, \eta$ are the weighting parameters that balance the three terms. The different points of the rate-semantic-perceptual curve can be obtained by changing the value of $\lambda$.

\subsection{The RL-based Semantic Bit Allocation Model}

Different semantic concepts matter differently in downstream analysis tasks, and only the task-related semantic information needs to be transmitted with high precision. The critical semantic concepts should be focused on and then encoded precisely without semantic loss, while the precision requirements for other objects can be relaxed. Taking the street scene dataset Cityscapes as an example, the object detection task mainly concentrates on the vehicle, pedestrian, and traffic sign, etc., while the background such as sky and vegetation does not affect the core meaning of the image.  Besides, a few bits can already well represent sky and vegetation, since their texture is simple and regular.


\begin{figure*}[t]
\centering
\includegraphics[width=1.0\linewidth]{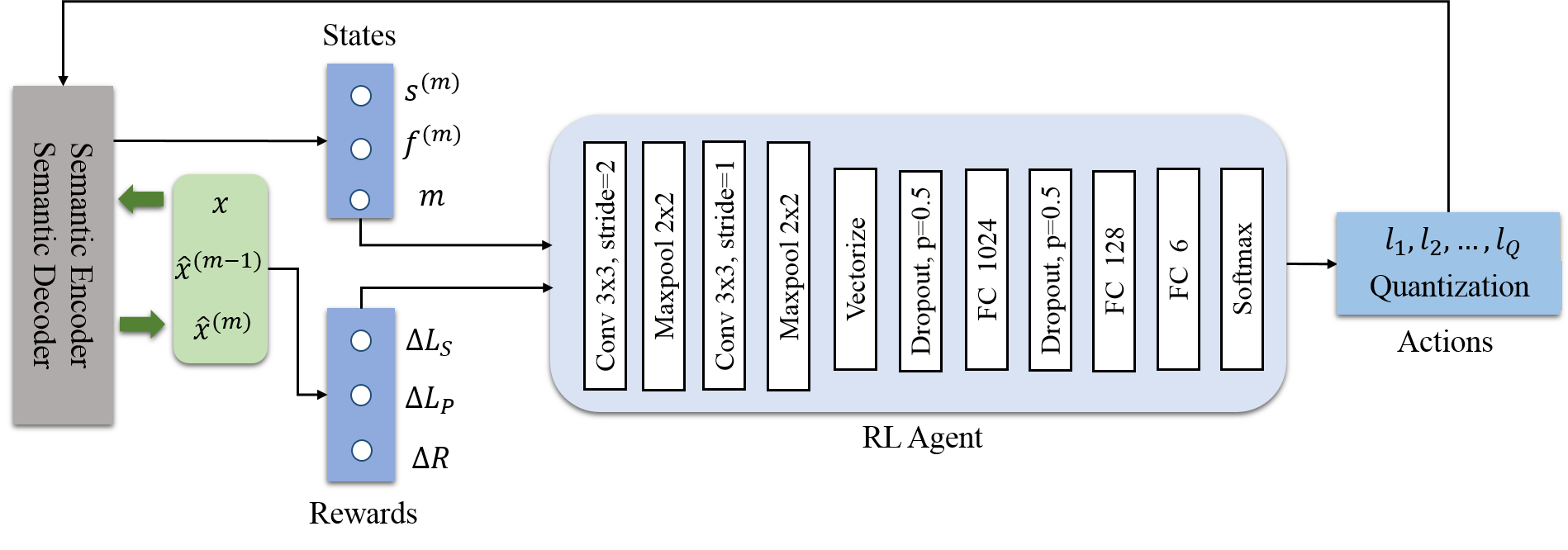}
\label{a}
\caption{The architecture of the RL-based Semantic Bit Allocation Model.}
\label{policy}
\end{figure*}

In order to locate the critical semantic concepts and assign appropriate precision requirement, 
we propose an RL-based semantic bit allocation model. After all the semantic concepts are obtained by the semantic encoder, the coding process is modeled as a Markov Decision Process (MDP), where each semantic concept is encoded and decoded sequentially with appropriate bit rate in the label index order from $m=1$ to $m=M$.  The RL agent adaptively assigns the quantization level for each semantic concept by balancing the bit cost and reconstruction performance. Accordingly, we can integrated the semantic fidelity and perceptual quality metrics into the semantic coding optimization. Therefore, the more informative and visually salient objects are represented with higher bitrates while the irrelevant regions are represented with lower bitrates. When the training of the RL agent is completed, the whole process of the 
quantization decision is off-policy, which can be processed in
parallel with the encoder.

The RL framework typically consists of a state space $\mathbb{S}$ that includes all the possible information that the agent can observe from the environment. The state in the $m$th step is $state^{(m)}=\{\pmb{f}^{(m)},\pmb{s}^{(m)}, m\}$, and we denote $\hat{\pmb{x}}^{(m-1)}$ as the reconstructed image from the former step. The action space $\mathbb{A}$ is defined as a set of quantization levels $action^{(m)}\in\{l_1,l_2,...,l_Q\}$ assigned for the semantic concepts, where $Q$ is the total number of levels and a higher quantization level leads to a higher bit rate and finer details.  We define the behavior of the agent as a policy $\pi: \mathbb{S}\times \mathbb{A} \rightarrow [0,1]$ that maps states to a distribution of actions, denoted as $\pi(action^{(m)} | state^{(m)})$.  Specifically, $\pi$ can be implemented by a neural network parameterized by $\theta$. The intermediate reward $r^{(m+1)}$ is received after the agent $\pi$  taking action $action^{(m)}$ at sate $state^{(m)}$ by evaluating the performance of the current reconstructed image $\hat{\pmb{x}}^{(m)}$. The transition function $trans(state^{(m)},action^{(m)})$ predicts the next state given the current state and the action. In this work, the evolution from state $state^{(m)}$ to $state^{(m+1)}$ is deterministic, and thus $p(state^{(m+1)})=trans(state^{(m)},action^{(m)})=1$.

As shown in Fig. \ref{policy}, the detailed procedure of the agent $\pi$ is as follows. The current state $state^{(m)}$ is fed into the agent to learn the policy $\pi$ that reveals the semantic importance and accordingly produces an action $action^{(m)}$ that adaptively quantizes $\pmb{C}^{(m)}$. Next, the semantic decoder takes the received bitstream and the selected quantization level as input to produce the current reconstructed image $\pmb{\hat{x}}^{(m)}$. The immediate reward $r^{(m+1)}$ is defined as the quality increase from $\hat{\pmb{x}}^{(m)}$ to $\hat{\pmb{x}}^{(m+1)}$ in terms of the increase of semantic similarity, perceptual quality, as well as the decrease in bitrate $R$. Note that compared to $\pmb{\hat{x}}^{(m-1)}$, the current $\pmb{\hat{x}}^{(m)}$ is updated in the region of the particular semantic concept $\pmb{C}^{(m)}$, while other region maintains unchanged. The bitrate $R$ of $\pmb{\hat{x}}^{(m)}$ is denoted as $\psi(\pmb{\hat{x}}^{(m)})$ and computed as the bitrate summation of the so far processed category-specific features $\pmb{f}^{(m)}$'s as well as the segmentation map $\pmb{s}$. The detailed bitrate calculation method is expressed in Section III. A.

We next design the architecture of RL agent $\pi$. Firstly, the state $state^{(m)}$ is fed into two consecutive convolutional layers, each of which is followed by the ReLU activation and max-pooling operation. Next, the output of the convolutional layer is flattened into a vector and then fed into three FC layers to yield an $M\times 1$ dimensional logits. Note that the dropout layer and ReLU activation are adopted to avoid over-fitting. Thirdly, the subsequent softmax function returns the predicted probability of each quantization level, and the action $action^{(m)}$ is sampled from the predicted probability distribution.

   At the initialization step, we set the coarsest quantization level $l_1$ for the whole image  to obtain an initial reconstruction $\pmb{\hat{x}}^{(0)}$ with the lowest bit rate $\psi(\pmb{\hat{x}}^{(0)})$. Then, both of the two images $\pmb{x}$ and $\pmb{\hat{x}}^{(0)}$ are fed into the downstream analysis network $\mathcal{H}$ (such as PSPNet) and the VGG network to measure the semantic loss and perceptual loss. At the next step, the state $state^{(1)}=\{\pmb{f}^{(1)}, \pmb{s}^{(1)}, m=1\}$ is take into consideration, and the action $action^{(1)}$ is produced by the agent to select a proper quantization level. The current reconstructed image $\hat{\pmb{x}}^{(1)}$ is obtained by the semantic decoder and sequentially the reconstruction performance is evaluated. 
 At step $m$, the rate-semantic-perceptual loss can be written as
\begin{equation} \label{reward}
    L^{(m)} =  \lambda[\psi(\pmb{\hat{x}}^{(m)})-\psi(\pmb{\hat{x}}^{(0)})] +  f_{degrade}(\mathcal{H}(\pmb{x}),\mathcal{H}(\pmb{\hat{x}}^{(m)}) ) + \eta\|\phi(\pmb{x})-\phi(\pmb{\hat{x}}^{(m)})\|_2.
\end{equation}
At step $m+1$, the $(m+1)$th semantic concept $\textbf{C}^{(m+1)}$ on the current reconstruction $\hat{\pmb{x}}^{(m+1)}$ is updated and we will accordingly obtain a renewed rate-semantic-perceptual loss $L^{(m+1)}$. The intermediate reward could be written as
\begin{equation}
    r^{(m+1)} =  L^{(m)} - L^{(m+1)}= \lambda \triangle R + \triangle L_S + \eta \triangle L_P,
\end{equation}
where $\triangle L_S, \triangle L_P$ and $\triangle R$ represent the difference in the semantic loss, the perceptual loss, and the bit rate before and after coding.

The agent should be driven to maximize the cumulated reward in the whole episode, which is the optimal goal of the MDP. Denote $\gamma$ as the discounted factor, the discounted cumulated reward for a full episode can be written as 
\begin{equation}\label{eqR}
G=\sum_{m=1}^M \gamma^m r^{(m+1)}.
\end{equation}
The objective function for such a learning process can be formulated as maximizing the expected $G$ for all trajectories. Denote $T$ as a complete trajectory $(state^{(1)}, action^{(1)}, r^{(2)}, state^{(2)}, action^{(2)},\\r^{(3)},...,$ $state^{(M)}, action^{(M)}, r^{(M+1)})$, and $T_{\pi}$ as the trajectory distribution. Denote the total number of sampled trajectories as $N$. Then the objective function can be written as
\begin{equation}
\begin{aligned}
    J(\pi) &= \mathbb{E}_{T_{\pi}}G=\sum_N T_{\pi} G.
\end{aligned}
\end{equation}
The optimal policy $\pi$ can be obtained by performing gradient ascend method on the sampled trajectories. The derivative of $J_{\pi}$ can be calculated as
\begin{equation}\label{J_deri}
\begin{aligned}
        \triangledown_{\theta} J_{\pi} = \sum_{N}G \triangledown_{\theta} T_{\pi}
        = \sum_{N} G T_{\pi} \frac{\triangledown_{\theta} T_{\pi}}{T_{\pi}}
        =\sum_{N}  T_{\pi} G\triangledown_{\theta} \log T_{\pi}.
\end{aligned}
\end{equation} 
 
Since $trans(state^{(m)},action^{(m)})$ is deterministic, we may further expand $T_{\pi}$ as
\begin{equation}\label{T}
\begin{aligned}
    T_{\pi}&=p(state^{(1)})\prod_{m=1}^{M} \pi(action^{(m)}|state^{(m)})trans(state^{(m)},action^{(m)})\\
    &=p(state^{(1)})\prod_{m=1}^{M} \pi(action^{(m)}|state^{(m)}),
    \end{aligned}
\end{equation}
where $p(state^{(1)})$ is the probability of observing $state^{(1)}$. For computational efficiency, we calculate the derivative of the logarithm form of formula \eqref{T} as
\begin{equation}\label{deriT}
\begin{aligned}
   \triangledown_{\theta} \log T_{\pi} &=  \triangledown_{\theta} \log p(state^{(1)})+ \sum_{m=1}^{M} \triangledown_{\theta}  \log \pi(action^{(m)}|state^{(m)})\\
  & =  \sum_{m=1}^{M} \triangledown_{\theta}  \log \pi(action^{(m)}|state^{(m)}),
   \end{aligned}
\end{equation}
where the first term $\log p(state^{(1)})$ is irrelevant to $\theta$ and thus the derivative is zero. According to \eqref{deriT}, equation \eqref{J_deri} can be rewritten as
\begin{equation} \label{rewritten}
    \triangledown_{\theta} J_{\pi} =
    \mathbb{E}_{T_{\pi}}[ G \sum_{m=1}^M \triangledown_{\theta} \log \pi(action^{(m)}|state^{(m)}) ].
\end{equation}
In practice, we can use a one-time Monte-Carlo rollout to sample a trajectory. Therefore, equation \eqref{rewritten} can be written as:
\begin{equation}\label{derivative}
    \triangledown_{\theta}  J_{\pi}= G \sum_{m=1}^M \triangledown_{\theta} \log \pi(action^{(m)}|state^{(m)}).
\end{equation}
Then, the parameters of the RL agent $\pi$ is updated as
\begin{equation}\label{update}
    \theta \leftarrow \theta + \alpha \triangledown_{\theta} J_{\pi},
\end{equation}
where $\alpha$ is the learning rate.

%

\section{Semantic Decoder and Training Details}
\begin{figure*}[t]
\centering
\includegraphics[width=1.0\linewidth]{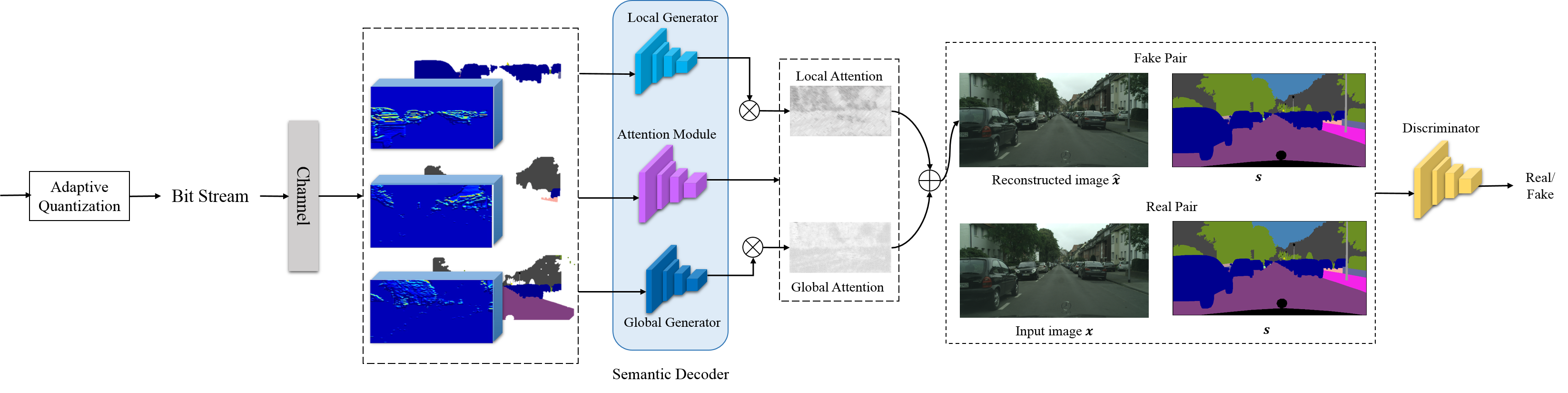}
\caption{The architecture of the semantic decoder.}
\label{traincoder}
\end{figure*}

As shown in Fig. \ref{traincoder}, we design the soft quantization, entropy coding, semantic decoder and the training details in this section.

\subsection{Soft Quantization and Entropy Coding}

The RL agent $\pi$ selects the action $a^{(m)}$ for each $\pmb{f}^{(m)}$ to assign bits adaptively and concentrate on semantic important regions.  We adopt the scalar variant for the quantization approach and obtain the discrete $\pmb{\hat{f}}^{(m)}$ as
  \begin{equation}
    \pmb{\hat{f}}^{(m)} := Quantize(\pmb{f}^{(m)}, a^{(m)}),
\end{equation}
where each entry of $\pmb{\hat{f}}_m$ at coordinate $(k,i,j)$ can be computed by the nearest neighbor assignment method. The process can be written as 
\begin{equation}\label{quantization}
    \pmb{\hat{f}}_m(k,i,j)=\argmin_t \| \pmb{f}^{(m)}(k,i,j) - l_m\|,
\end{equation}
where $l_m\in \{l_1, l_2,...,l_Q\}$ is the quantization center.

A major problem in quantization is that the gradients of \eqref{quantization} are zeros almost everywhere, which makes gradient descent-based optimization ineffective in the end-to-end communication system. To address the zero gradient problem and approximate formulation \eqref{quantization},  we introduce a differentiable soft quantization with continuous functions \cite{2019Differentiable}
\begin{equation}\label{soft}
    \pmb{\tilde{f}}^{(m)}(i,j,k) = \sum_{m=1}^{Q} \frac{exp(-\sigma \|\pmb{f}^{(m)}(k,i,j)  - l_m\|)}{\sum_{t=1}^{Q} exp(-\sigma \|\pmb{f}^{(m)}(k,i,j)  - l_t\|)}l_m,
\end{equation}
where $\sigma$ is a hyperparameter relating to the softness of the quantization.
Note that the quantization formulation \eqref{quantization} is adopted in the forward pass, while the differentiable soft quantization formulation \eqref{soft} is adopted in the backward pass.

It is already known that the entropy coding can reduce the expected bit length by assigning a short code for frequently occurred codeword while assigning a long code for others. We adopt Huffman coding as the entropy coding, where the codeword probability is predicted by building and maintaining a frequency table.  In this case, $\pmb{\hat{f}}^{(m)}$ is transformed into the variable-length binary code $\pmb{r}^{(m)}$ of length $\ell(\pmb{r}^{(m)})$ as
\begin{equation}\label{huffman}
    \pmb{r}^{(m)}:= Huffman(\pmb{\hat{f}}^{(m)})\in \{0,1\}^{\ell(\pmb{r}^{(m)})}.
\end{equation}


Besides, the semantic label map $\pmb{s}$ is losslessly coded in vector graphic format, and the binary code is denoted as $\pmb{r}_s$. The coded bitstream of the proposed RL-ASC includes two parts: (i) the binary codes $\pmb{r}_s$ with fixed length $\ell(\pmb{r}_{\pmb{s}})$; (ii) the binary codes $\pmb{r}^{(m)}$ with variable length $\ell(\pmb{r}_m)$ that is adjusted by $a^{(m)}$.
We thereby define the rate  of the reconstructed image $\hat{\pmb{x}}^{(m)}$ at step $m$ as the total bits of all the binary codes $\pmb{r}^{(m)}$ and $\pmb{r}_s$, which can be expressed as
\begin{equation}
    \psi(\hat{\pmb{x}}^{(m)}) = \ell(\pmb{r}_{\pmb{s}}) + \sum_{m=1}^M(\ell(\pmb{r}^{(m)})).
    \label{rate}
\end{equation}

Then, the binary code $\pmb{r}^{(m)}$ and $\pmb{r}_s$ are transmitted through the wireless channel and arrive at the receiver. Unless notified, we assume the wireless channel is lossless since the paper mainly focuses on image semantic encoding and semantic decoding. Therefore, the receiver could achieve perfect $\pmb{\hat{f}}^{(m)}$ and $\pmb{s}$. Benefiting from the incredible semantic extraction and bit allocation performance, the essential semantics are appropriately preserved and transmitted.


\subsection{Generative Semantic Decoder} 

 In this section, the received bit streams are decoded into semantic concepts and then combined as the reconstructed image. Some learning-based codecs are based on convolutional neural networks (CNNs) \cite{ICLR,2018Nips,20193507365934,Hu2020CoarsetoFineHM}, while others on recurrent neural networks (RNNs) \cite{20191106642196,20181304961211}.
 Recent
learning-based codecs \cite{20204409437480,2021Variable} are competitive with or
even superior to the classical codecs.  The work \cite{20204409437480} proposed to use discretized Gaussian Mixture Likelihoods to
parameterize the distributions of latent codes, which can achieve a more accurate and flexible entropy model. 
The work \cite{2021Variable} propose a versatile deep image compression network based on Spatial Feature Transform, which enables task-aware image compression for various tasks, e.g., classification, with variable rates. However, all of the aforementioned works optimizes the rate-distortion (RD) performance and cannot meet the requirements of ISC.

\begin{figure*}[t]
\centering
\includegraphics[width=0.9\linewidth]{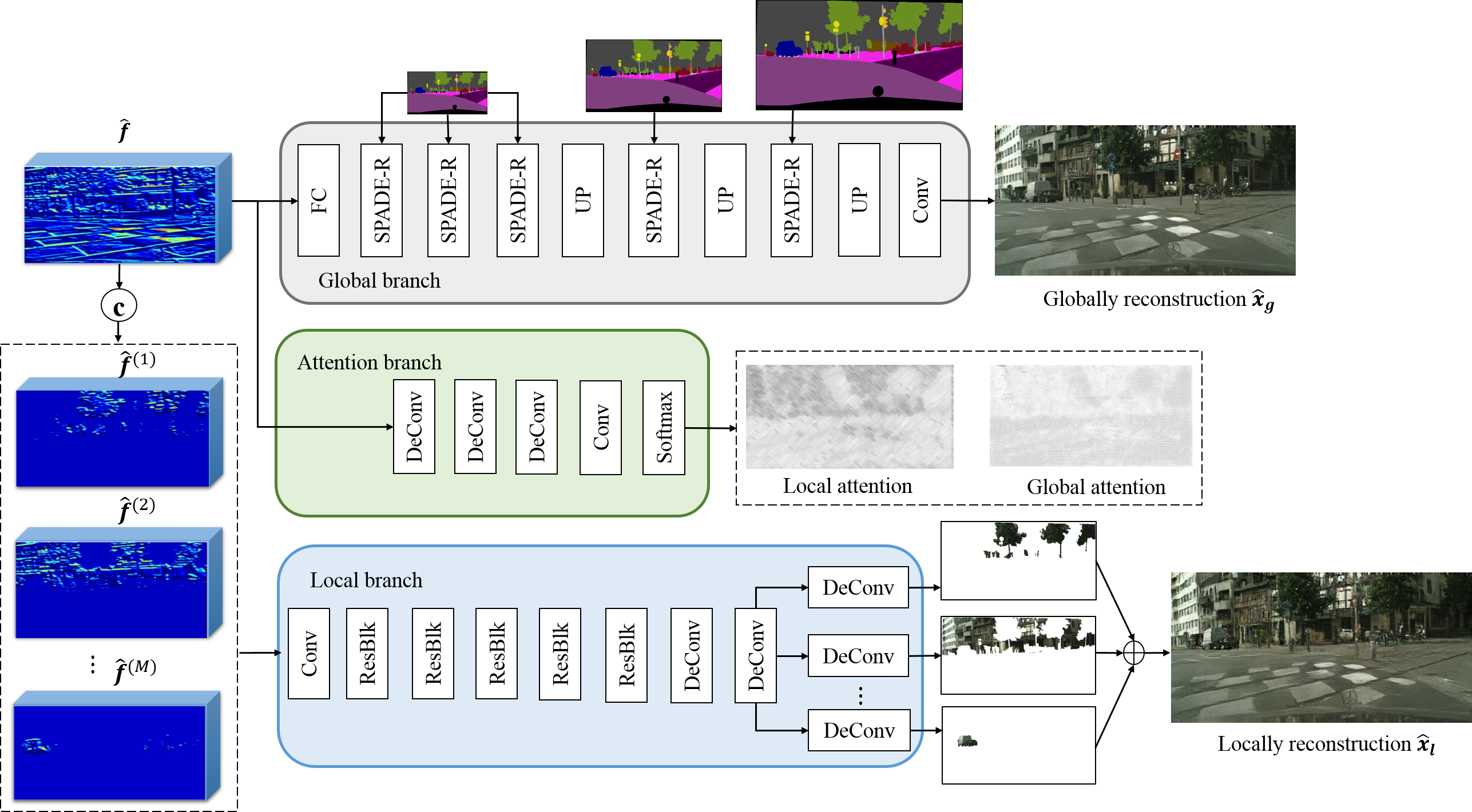}
\caption{The architecture of the generative semantic decoder.}
\label{LG}
\end{figure*}

 Different from conventional methods that reduce pixel errors, the proposed semantic decoder should be tolerant of pixel errors to some extent. GANs \cite{goodfellow2014generative} has shown impressive ability in synthesizing high-resolution images. Instead of forcing the generated images to be exactly the same as the ground truth image, GANs attempt to approximate the intractable distributions of the real image dataset. Therefore, the generated image maintains the underlying meaning of the ground truth image while the two images may differ in pixel values. Therefore, GANs fits the spirit of the ISC and are adopted as the semantic decoder in the proposed method. 

Adversarial training of the GANs has shown  advantage in reducing compression artifacts in
 deep compression systems  \cite{Theis,pmlr-v70-rippel17a,ICLR,Akbari}. Inspired by existing methods, we adopt conditional GANs in the semantic decoder that translates the semantic concepts into the reconstructed image. The reconstructed image should be natural and similar to the input image.
The naturalness can be measured by the adversarial loss that indicates the probability distributions divergence of real and reconstructed images, while the visual similarity could be measured by the aforementioned perceptual loss that indicates the distance in feature space  \cite{Blau}. 

The detailed architecture of the generative semantic decoder is illustrated in Fig. \ref{LG}, which consists of a local generator $\mathcal{G}_l$, a global generator $\mathcal{G}_g$, and an attention module.

\textbf{Local Generator $\mathcal{G}_l$:} The quantity and spatial occupation of different categories are imbalanced in the training dataset.  It is extremely difficult to generate small object classes and texture details since the dataset are dominated by  frequently occurred or large object classes. To avoid the interference from other classes, we propose a novel local image generator $\mathcal{G}^l$ to separately reconstruct the class-specific objects $\pmb{y}^{(m)}\in \mathbb{R}^{H \times W}$ from the quantized class-specific feature maps $\pmb{\hat{f}}^{(m)}$ as
\begin{equation}
    \pmb{y}^{(m)} = \mathcal{G}^l(\pmb{\hat{f}}^{(m)}).
\end{equation}
The locally reconstructed image $\pmb{\hat{x}}_l\in \mathbb{R}^{H\times W}$ is obtained by element-wise addition of all the individual $\pmb{y}^{(m)}$'s as
\begin{equation}
    \pmb{\hat{x}}_l = \pmb{y}^{(1)} \oplus \pmb{y}^{(2)} \oplus \cdot\cdot\cdot \oplus \pmb{y}^{(M)}.
\end{equation}

As shown in Fig. \ref{LG}, we feed $M$ feature maps $\pmb{\hat{f}}^{(m)}$ into a convolution layer and five consecutive residual blocks, and the outputs are then up-scaled by two consecutive deconvolutional layers. Then, the up-scaled feature maps of $M$ semantic concepts are separately fed into $M$ corresponding deconvolutional layers to produce $\pmb{y}^{(m)}$. Each deconvolutional layer has independent network parameters and is able to effectively preserve the class-specific features with rich details.

\textbf{Global Generator $\mathcal{G}_g$:} 
We attempt to capture the global structure information as well as the spatial layout by global generation. Inspired from GauGAN \cite{Park}, the well designed $\mathcal{G}_g$ adopts spatially adaptive normalization (SPADE) to fuse the spatial layout information from $\pmb{s}$ into the feature maps $\pmb{\hat{f}}$. The SPADE modulates the layer activation in accordance with $\pmb{s}$ to guide the reconstruction of different categories. The globally reconstructed image $\pmb{\hat{x}}_g$ can be obtained as
\begin{equation}
    \pmb{\hat{x}}_g = \mathcal{G}_g(\pmb{\hat{f}},\pmb{s}).
\end{equation}

Particularly, we feed $\pmb{\hat{f}}$ into three residual blocks with SPADE (SPADE-R), the output of which is upsampled three times to yield the global reconstruction $\pmb{\hat{x}}_g$. 

\textbf{Attention Module:} In order to better combine the outputs of $\mathcal{G}_l$ and $\mathcal{G}_g$, we further propose an attention module to learn the local weight matrix $\pmb{W}_l\in\mathbb{R}^{H\times W}$ and global weight matrix $\pmb{W}_g\in\mathbb{R}^{H\times W}$. The feature map $\pmb{\hat{f}}$ is fed into the attention module that includes three consecutive deconvolutional layers and a convolutional layer. The output of the convolutional layer has two channels that correspond to the two weight matrices and is further normalized by a softmax layer in channel-wise. Therefore, we obtain $\pmb{W}_g$ and $\pmb{W}_l$ whose values are in the range of (0,1) and meet the condition $\pmb{W}_l(i,j)+\pmb{W}_g(i,j)=1$ in each location $(i,j)$. The final output image $\pmb{\hat{x}}$ can be written as element-wise summation of weighted local reconstruction $\pmb{\hat{x}}_l$ and global reconstruction $\pmb{\hat{x}}_g$:
\begin{equation}\label{output}
   \pmb{\hat{x}}= (\pmb{W}_l\odot \pmb{\hat{x}}_l) \oplus (\pmb{W}_g\odot \pmb{\hat{x}}_g).
\end{equation}

We adopt a multi-scale patch-discriminator $\mathcal{D}$ to identify the real image $\pmb{x}$ and the final output image $\pmb{\hat{x}}$. The generator $\mathcal{G}_l, \mathcal{G}_g$ are trained alternatively with the discriminator $\mathcal{D}$. Denote $\mathcal{D}(\pmb{x},\pmb{s})$ or $\mathcal{D}(\pmb{\hat{x}},\pmb{s})$ as the predicted probability that the image pair $(\pmb{x},\pmb{s})$ or $(\pmb{\hat{x}},\pmb{s})$ are from real samples. The discriminator $\mathcal{D}$ learns to distinguish real image pairs from fake ones by maximizing $\mathcal{D}(\pmb{x},\pmb{s})$ and minimizing $\mathcal{D}(\pmb{\hat{x}},\pmb{s})$. The generators $\mathcal{G}_l$ and $\mathcal{G}_g$ learn to fool $\mathcal{D}$ by minimizing  $(-\mathcal{D}(\pmb{\hat{x}},\pmb{s}))$. We adopt the hinge loss for GANs, whose objective function is
\begin{equation}\label{objective}
\begin{aligned}
     \min_{\mathcal{D}} L_{\mathcal{D}} &= \mathbb{E}[\max(0,1-\mathcal{D}(\pmb{x},\pmb{s}))] + \mathbb{E}[\max(0, 1+\mathcal{D}(\pmb{\hat{x}},\pmb{s}))],\\
     \min_{\mathcal{G}_l, \mathcal{G}_g} L_{\mathcal{G}} &= -\mathbb{E}[\mathcal{D}(\pmb{\hat{x}},\pmb{s})].
\end{aligned}
\end{equation}
Note that the hinge loss $L_\mathcal{D}$ is used for ``maximum-margin" classification, which penalizes the positive samples of $\mathcal{D}(\pmb{x},\pmb{s})<1$ and the negative samples of $\mathcal{D}(\pmb{\hat{x}},\pmb{s})>-1$. Only the samples that are not classified properly will impact the gradient update of $\mathcal{G}_l$, $\mathcal{G}_g$ and $\mathcal{D}$. The formulation \eqref{objective} forces the probability distribution of $\pmb{\hat{x}}$ to approximate that of the training image set.  If the feature extraction network $\mathcal{E}$ cannot afford to store the exact detail in $\pmb{x}$, then $\mathcal{G}^l$ and $\mathcal{G}^g$ are able to synthesize the detail to satisfy natural image distribution instead of showing blocky and blurry effects. Equilibrium will be achieved when $\mathcal{D}$ classifies the reconstructed image $\pmb{\hat{x}}$ as real.

The final reconstruction $\pmb{\hat{x}}$ should not only be consistent with the natural image distribution, which coincides with the target of traditional GANs, but also meet the requirements of semantic communications and recover the semantic concepts of a specific input image $\pmb{x}$. We adopt the perceptual loss \eqref{distortion2} to guide the semantic reconstruction process and ensure $\pmb{\hat{x}}$ approximates the feature space of $\pmb{x}$.

The proposed semantic encoder and semantic decoder are trained jointly, and the objective function can be formulated as the weighted combination of adversarial loss $L_\mathcal{D}$, perceptual loss $L_P$, and feature classification loss $L_C$ as
\begin{equation}\label{loss}
     \min_{\mathcal{E},\mathcal{G},\mathcal{D}} L_{\mathcal{D}} + \lambda_1 L_P + \lambda_2 L_C,
 \end{equation}
 where $\lambda_1$ and $\lambda_2$ are the weighting parameters. Due to the adequately preserved semantic information, the proposed generative semantic decoder produces more incredible reconstructions compared with the leading image generation models \cite{Park,Agustsson,Akbari}. Note that the work \cite{Agustsson} performs poor in preserving appearance features, and the reconstructed image deviates apparently from the source image. The work \cite{Park} can merely generate a general image from the distribution of the training dataset without the ability to control the appearance of certain objects. 

\subsection{Training Algorithm}
We next propose a three-stage training algorithm to ensure the semantic encoder, the semantic decoder, and the RL-based semantic bit allocation model function properly. Benefit from the differentiable soft quantization, the whole system can be trained with an ADAM solver \cite{kingma2014adam}, where the momentum term of ADAM is set to be $\beta_1=0.5$ and $\beta_2=0.999$. The whole process is to first conduct stage \uppercase\expandafter{\romannumeral1} and train the semantic encoder and decoder in an end-to-end manner, while taking equal quantization for every semantic concepts with the highest quantization level. Then, with the pretrained semantic encoder and decoder being fixed, we conduct stage \uppercase\expandafter{\romannumeral2} to train the semantic bit allocation model, a.k.a, the RL agent $\pi$. In stage III, the semantic encoder, semantic decoder, and RL agent $\pi$ are finetuned together.

\begin{algorithm}[t]

\caption{Training the semantic encoder and semantic decoder}
\small
\label{alg2}
\LinesNumbered
\KwIn{Image dataset $\pmb{X}$, pretrained $\mathcal{F}$, training epoch $E_p=70$, learning rate $\alpha=2e^{-4}$, parameters $\lambda_1$ and $\lambda_2$, batch size $b$, total categories of semantic concepts $M$, and fixed quantization level $a^{(M)}$.}
\KwOut{Neural network parameter for $\mathcal{G}_l$, $\mathcal{G}_g$, $\mathcal{D}$, and $\mathcal{E}$.}
\For{epoch=1:$E_p$} {
Sample a batch of image $\{\pmb{x}_1,...,\pmb{x}_b\}\in \pmb{X}$;\\
Obtain $M$ feature maps $\pmb{\hat{f}}_1^{(m)},...,\pmb{\hat{f}}_b^{(m)}$ and $\{\pmb{s}_1, ..., \pmb{s}_b\}$ for the image batch by $\mathcal{E}$ and $\mathcal{F}$;\\
Quantize $M$ feature maps $\pmb{\hat{f}}_1^{(m)},...,\pmb{\hat{f}}_b^{(m)}$ with quantization level $a^{(M)}$ by formulation \eqref{soft};\\
The bit streams are transmitted to the receiver through channel;\\
Obtain the final output image $\{\pmb{\hat{x}}_1, ..., \pmb{\hat{x}}_b\}$ by $\mathcal{G}_l$ and $\mathcal{G}_g$ by formulation \eqref{output};\\
Update the discriminator $\mathcal{D}$ by descending its stochastic gradient:
$\nabla_{\theta_d} \frac{1}{b} \sum_{j=1}^b \{[\max(0,1-\mathcal{D}(\pmb{x}_j,\pmb{s}_j))] + [\max(0, 1+\mathcal{D}(\pmb{\hat{x}}_j,\pmb{s}_j))]\}$.\\
Update the generators $\mathcal{G}_l$, $\mathcal{G}_g$  and $\mathcal{E}$ by descending their stochastic gradient: $\nabla_{\theta_{g,e}} \frac{1}{b} \sum_{j=1}^b [-\mathcal{D}(\pmb{\hat{x}}_j,\pmb{s}_j)+\lambda_1 L_{P} + \lambda_2 L_C]$.
}
Return the network $\mathcal{G}_l, \mathcal{G}_g, \mathcal{D}$, and $\mathcal{E}$.

\end{algorithm}

\begin{algorithm}[t]
\caption{Training the RL-based semantic bit allocation model}
\small
\label{alg}
\LinesNumbered
\KwIn{Image dataset $\pmb{X}$, training epoch $E_p=5$, learning rate $\alpha=1e^{-5}$, parameters $\eta=10$ and $\lambda=1$, batch size $b=1$, total categories of semantic concepts $M$, quantization levels $\mathbb{A}=\{l_1,l_2,...,l_Q\}$, discounted factor $\gamma=0.99$, the pre-trained semantic encoder and semantic decoder.}
\KwOut{Neural network parameter $\theta$.}
\For{epoch=1:$E_p$}{
Sample a batch of image data $\pmb{x}\in \pmb{X}$.\\
Initialization: Set all the $M$ semantic concepts $\textbf{C}^{(m)}$'s with the coarsest quantization level $a^{(0)}=l_1$ and obtain the bitrate $\psi(\pmb{x}^{(0)})$ by \eqref{rate}.\\
\For{m=1:M}{
\textbf{Encoding}: The semantic encoder extracts $\textbf{C}^{(m)}$ from $\pmb{x}$;\\
\textbf{Quantization}: The policy $\pi(a^{(m)}|s^{(m)})$ selects an action $a^{(m)}\in \mathbb{A}$ for $\textbf{C}^{(m)}$; \\
\textbf{Entropy Coding}: Conduct Huffman coding by formulation \eqref{huffman} and obtain bitrate $\psi(\pmb{x}^{(m)})$;\\
\textbf{Transmitting}: The bit streams are transmitted to the receiver through channel;\\
\textbf{Decoding}: The semantic decoder reconstructs $\pmb{\hat{x}}^{(m)}$ by formulation \eqref{output};\\
\textbf{Reward}: Calculate intermediate reward $r^{(m+1)}$ by formulation \eqref{reward}. \\
}
Calculate discounted cumulated reward $G$ by formulation \eqref{eqR}.\\
Calculate the gradient of $J_{\theta}$ by formulation \eqref{derivative}.\\
Update network parameter $\theta$ by formulation \eqref{update}.\\
}
Return the network parameter $\theta$.
\end{algorithm}

\textbf{Stage \uppercase\expandafter{\romannumeral1} Training the semantic encoder and semantic decoder}: In this stage, the proposed RL-ASC learns to extract semantic concepts and then semantically reconstruct the image, laying the basis for training the RL agent $\pi$.  
We first initialize $\mathcal{E}$, $\mathcal{G}_l$, $\mathcal{G}_g$, and $\mathcal{D}$ but leave the RL agent $\pi$ out of this stage. We assume all the semantic concepts $\textbf{C}^{(m)}$'s in the input image $\pmb{x}$ are of the same importance, and the quantization precision is set to be the highest level for each concept. Hence, the lowest information loss is obtained after quantization and the semantic decoder can produce the reconstruction $\pmb{\hat{x}}$ as fine as possible. The training algorithm is summarized in Algorithm \ref{alg2}.

Alternate training is applied between one gradient descent step on $\mathcal{D}$,  then one step on $\mathcal{G}_l$, $\mathcal{G}_g$, and $\mathcal{E}$. The training takes place in a loop, where the adversarial loss $L_\mathcal{D}$, the perceptual loss $L_P$, and the feature classification loss $L_C$ are minimized as in \eqref{loss}. Firstly, we update $\mathcal{D}$ by real training images and the reconstructed images. Next, with $\mathcal{D}$ being set as non-trainable, we feed $\pmb{\hat{f}}^{(m)}$ and $\pmb{s}$ to both $\mathcal{G}_l$ and $\mathcal{G}_g$ to produce final output image $\pmb{\hat{x}}$, and then we adopt $\mathcal{D}$ to identify $\pmb{\hat{x}}$ from real samples. Once the discrepancy between reconstructed and real training images is obtained, the parameters of $\mathcal{E}$, $\mathcal{G}_l$ and $\mathcal{G}_g$ can be updated by back-propagation.  The semantic encoder and semantic decoder are trained with learning rate $\alpha=2e^{-4}$ and  training epoch $E_p=70$. 

\textbf{Stage \uppercase\expandafter{\romannumeral2}: Training stage for semantic bit allocation model:} In this stage, we fix $\mathcal{E}$, $\mathcal{G}_l$, $\mathcal{G}_g$, and $\mathcal{D}$ obtained from stage \uppercase\expandafter{\romannumeral1}, and evoke a randomly initialized RL agent $\pi$ to be trained with reinforcement learning. Specifically, after selecting a quantization level $a^{(m)}$ for the current semantic concept $\textbf{C}^{(m)}$, the agent will receive a reward $r^{(m+1)}$ indicating whether this action is beneficial. The reward includes a decrease in perceptual loss, the semantic loss, and the rate after taking the action.  We train $\pi$ with an off-the-shelf policy gradient algorithm to maximize the cumulated discounted rewards $G$. The procedure for training the RL-based semantic bit allocation model is summarized in Algorithm \ref{alg}.

\textbf{Stage \uppercase\expandafter{\romannumeral3}: Fine tuning stage of the whole model:} In this stage, the semantic encoder $\mathcal{E}$, semantic decoder $\mathcal{G}_g, \mathcal{G}_l$ and RL agent $\pi$ are finetuned together.

\section{Experimental Results}
\subsection{Simulations Setup}
\subsubsection{Datasets}
We train and evaluate the RL-ASC model based on the scene parsing and instance segmentation of the Cityscapes dataset \cite{Cordts}. Cityscapes focuses on the semantic understanding of urban street scenes. It is collected from streetscapes in 50 different German cities and consists of 30 classes of objects. There are 2975 images in the training set and 500 images in the validation set; each being annotated with fine semantic labels. We downscale the images and semantic label maps to $256\times 512$ and conduct testing on the validation set. The classes or the semantic concepts in the Cityscapes dataset are listed as Table \ref{label}.

\begin{table}
\caption{The classes in the Cityscapes dataset.}
\label{label}
\centering
\begin{tabular}{c|c}
\hline
\hline
Group & Classes \\
\hline
flat & road · sidewalk · parking · rail track \\
\hline
human & person · rider \\
\hline
vehicle & car · truck · bus · on rails · motorcycle · bicycle · caravan · trailer*+\\
\hline
construction &	building · wall · fence · guard rail · bridge · tunnel\\
\hline
object &	pole · pole group · traffic sign · traffic light\\
\hline
nature &	vegetation · terrain\\
\hline
sky	 & sky\\
\hline
void &	ground · dynamic · static\\
\hline
\hline
\end{tabular}
\end{table}

\subsubsection{Baselines}
We compare the proposed RL-ASC with the engineered codecs BPG \cite{BPG}, JPEG2000 \cite{JPEG2000}, JPEG \cite{JPEG}, as well as the deep learning-based codecs DSSLIC \cite{Akbari} and HiFIC \cite{mentzer2020high}. Moreover, we finetuned the DSSLIC with the semantic loss in the semantic segmentation task for fair comparison, and the finetuned model is denoted as DSSLIC-finetuned. The bitrate of the engineered codecs \cite{BPG,JPEG2000,JPEG} is controlled by the quantization parameters (QP), where a larger QP means a higher compression ratio. We adopt the officially released version of these codecs and evaluate the performance in the range of bitrates 0 to 0.5 bpp. Particularly, BPG \cite{BPG} is the current state-of-the-art engineered image compression codec in terms of PSNR. DSSLIC \cite{Akbari} is a layered image compression, where the semantic label of the input image is encoded as the base layer of the bitstream, and the compact representation as well as the residual are encoded as the enhancement layer. The compression ratio of DSSLIC and DSSLIC-finetuned is adjusted by the QP of the enhancement layer. HiFIC \cite{mentzer2020high} combines the GANs with learned compression to achieve high fidelity generative lossy compression, and thus is able to obtain visually pleasing reconstructions that are perceptually similar to the input and operate in a broad range of bitrates. We evaluate the performance of \cite{mentzer2020high} by leveraging the pre-trained models at low bitrate (0.18bpp) and medium bitrate (0.33bpp). 

\subsubsection{Evaluation Metrics}
To measure the efficiency of the proposed RL-ASC, we evaluate the reconstruction performance from the semantic and perceptual perspectives. The most widely used quality metrics PSNR and SSIM are simple, shallow functions, and fail to account for many nuances of human perception. On the one hand, we evaluate the semantic loss in terms of the performance of downstream tasks such as object detection and semantic segmentation. We adopt mIoU as the objective metric and also illustrate the results of semantic segmentation and object detection as the subjective evaluation. The reconstructed image that suffers less loss in semantic information could yield higer mIoU value, and the maximum value of mIoU is 1.

On the other hand, the metrics Fréchet Inception distance score (FID) \cite{Heusel}, and Kernal-Inception distance (KID) \cite{Binkowski} (lower better) are consistent with human perception, and thus are employed to evaluate the distance of the reconstructed image and input image in deep feature space. Specifically,  KID \cite{Binkowski} and FID \cite{Heusel} measure the distribution divergence of the reconstructed images compared with real samples via the Inception network and are widely used to assess sample quality and diversity in the context of GANs.  Moreover, the perceptual performance can also be evaluated subjectively by user study, and a better compression approach can yield real, natural, and visual pleasant reconstructions even at a low bitrate. 

\subsubsection{Compression Modes}
We train three bitrate models of the proposed RL-ASC: the low bitrate (0.08 bpp), the medium bitrate (0.16 bpp), and the high bitrate (0.32 bpp). The bitrate is adjusted by the channels dimension $n$ of the feature map $\pmb{f}$ that is produced by the feature extraction network $\mathcal{E}$. We set $n=16,32,64$, $w=W/8$ and $h=H/8$, and therefore the dimensions of $\pmb{f}$ correspond to the three modes are $16 \times W/8 \times H/8$, $32 \times W/8 \times H/8$, and $64 \times W/8 \times H/8$, respectively. Also, the dimension of the downscaled semantic mask $\pmb{s}_d^{(m)}$ is $ W/8 \times H/8$. Additionally, we set $Q=6$, so that the RL agent $\pi$ can choose six quantization levels  for different semantic concepts, and higher quantization level means small distortion.

To evaluate the effect of the RL-based semantic bit allocation model, we conduct the ablation study by removing the RL agent $\pi$ from the proposed RL-ASC model. Such an ISC system lacks the adaptive bit allocation ability and thus is denoted as simplified RL-ASC. Specifically, simplified RL-ASC encodes each semantic concept with equal precision by the highest quantization level. The three bitrate models for simplified RL-ASC are low bitrate (0.11 bpp), medium bitrate (0.22 bpp), and high bitrate (0.44 bpp), respectively.

\subsection{Semantic Performance}
We compare the semantic performance of the proposed RL-ASC and the simplified RL-ASC with the baseline codecs at different bitrates. To validate the effectiveness of the proposed RL-ASC, we apply the proposed method in two downstream semantic tasks: semantic segmentation and object detection.

\subsubsection{Objective Quality}
\begin{figure*}[t]
\centering
\includegraphics[width=0.6\linewidth]{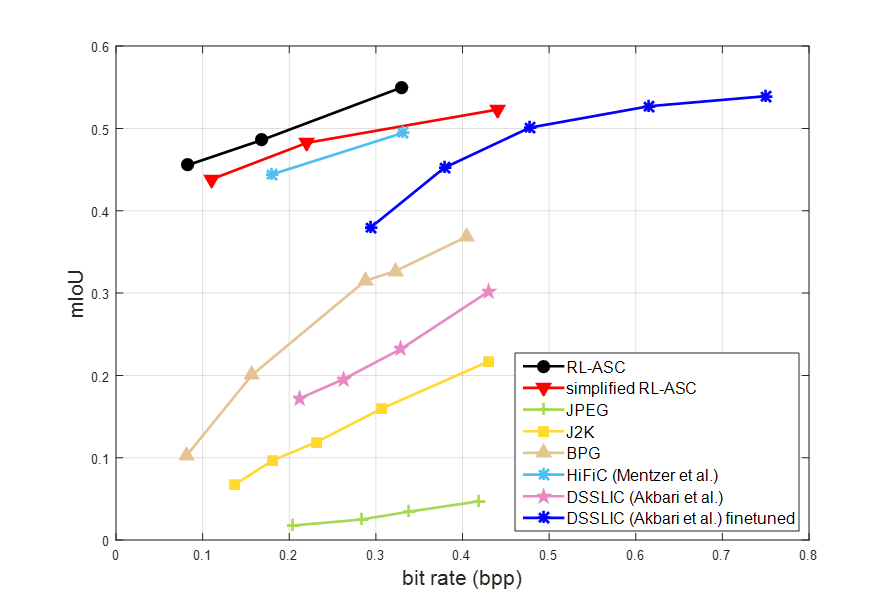}
\caption{The semantic performance in terms of mIoU of different image codecs on semantic segmentation task. The higher value means better performance.}
\label{mIoU}
\end{figure*}

The pretrained PSPNet \cite{Zhao} is adopted as the semantic segmentation model to obtain the semantic label $\pmb{\hat{s}}$ of the reconstructed image. The semantic fidelity can be measured by the consistency between $\pmb{\hat{s}}$ and the ground truth $\pmb{s}$ in terms of mIoU. The mIoU performance of different image codecs in different bitrates is shown in Fig. \ref{mIoU}. Cityscapes dataset only contains the ground truth label for semantic segmentation and lacks the ground truth label for other intelligent tasks such as object detection and image classification. Therefore, we cannot obtain the mAP metric of object detection task for objective measurement. 

\begin{table}
\caption{BD-mIoU and BD-rate relative to the baselines for semantic segmentation tasks.}
\label{BD-mIoU}
\centering
\begin{tabular}{c|c|c|c|c|c|c|c}
\hline
\hline
Metric & DSSLIC \cite{Akbari} &DSSLIC-finetune & HiFiC \cite{mentzer2020high} & BPG \cite{BPG} & J2K \cite{JPEG2000} & JPEG \cite{JPEG} & Simplified RL-ASC\\
\hline
 BD-mIoU & 0.278 & 0.145 & -0.004 & 0.227 & 0.343 & 0.452 & 0.0261\\
BD-rate & -97.713\%  &-60.769\% & 9.951\% & -89.682\% & -99.688\% & -100\% &-31.361\%\\
\hline
\hline
\end{tabular}
\end{table}

According to Fig. \ref{mIoU}, the semantic fidelity increases with higher bitrates. The proposed RL-ASC outperforms JPEG \cite{JPEG}, J2K \cite{JPEG2000}, BPG \cite{BPG}, DSSLIC \cite{Akbari} and DSSLIC-finetuned by a large margin in terms of mIoU, which validates the effectiveness of the proposed task-driven coding manner. The baseline HiFiC \cite{mentzer2020high} approximates the performance of the proposed RL-ASC, while it fails to achieve an extremely low bitrate ($<0.1$ bpp). The RL-ASC achieves higher mIoU performance compared to the simplified RL-ASC. Note that the DSSLIC finetuned by the semantic loss boosts the performance on the intelligent task in terms of mIoU.


We utilize the Bjontegaard metric \cite{Bjntegaard2001CalculationOA} to evaluate the coding efficiency of the proposed RL-ASC concerning the baselines. Inspired from \cite{9472999}, we propose BD-mIoU to consider the relative differences between two codecs under equal bitrate in task-related accuracy. BD-mIoU calculates the average mIoU difference between two rate-semantic curves over an interval and the BD-rate represents the average bitrate reduction under the equivalent task-related accuracy. As shown in TABLE. \ref{BD-mIoU}, the proposed RL-ASC method can achieve the same mIoU with more than 60\% bitrate savings on average compared with the deep learning based methods \cite{mentzer2020high, Akbari} and DSSLIC-finetuned. Under the same bit cost, the proposed RL-ASC method can remarkably improve the mIoU performance compared to the baselines and simplified RL-ASC.

\subsubsection{Subjective Quality}

\begin{figure*}[t]
\centering
\includegraphics[width=1.0\linewidth]{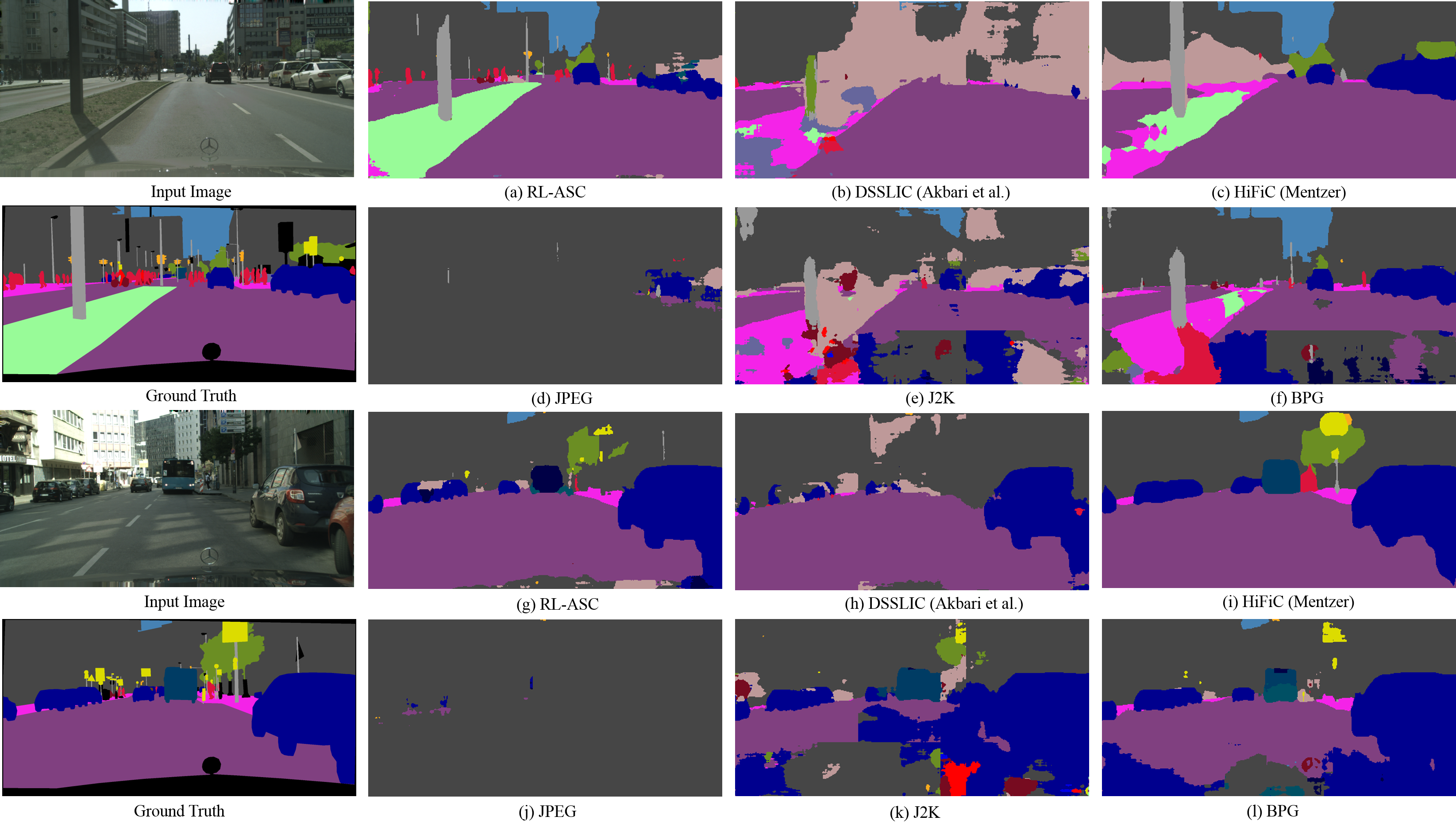}
\caption{Examples of image coding on downstream semantic segmentation. The first column contains the two randomly selected input image before coding and the corresponding ground truth labels.  The rest columns are the semantic label maps of decompressed images of different codecs under similar bitrate (0.33 bpp).}
\label{segmentation}
\end{figure*}
In this section, we visualize the semantic segmentation and object detection results of the decompressed images to validate the semantic fidelity subjectively. To ensure a fair comparison, the proposed RL-ASC encodes the image at a bitrate 0.33 bpp while the baselines encode the image at a bitrate equal to or higher than 0.33 bpp. Particularly, the MaskRCNN  \cite{he2018mask} pretrained on COCO dataset is adopted to detect objects on the reconstructed image. 

\begin{figure*}[t]
\centering
\includegraphics[width=1.0\linewidth]{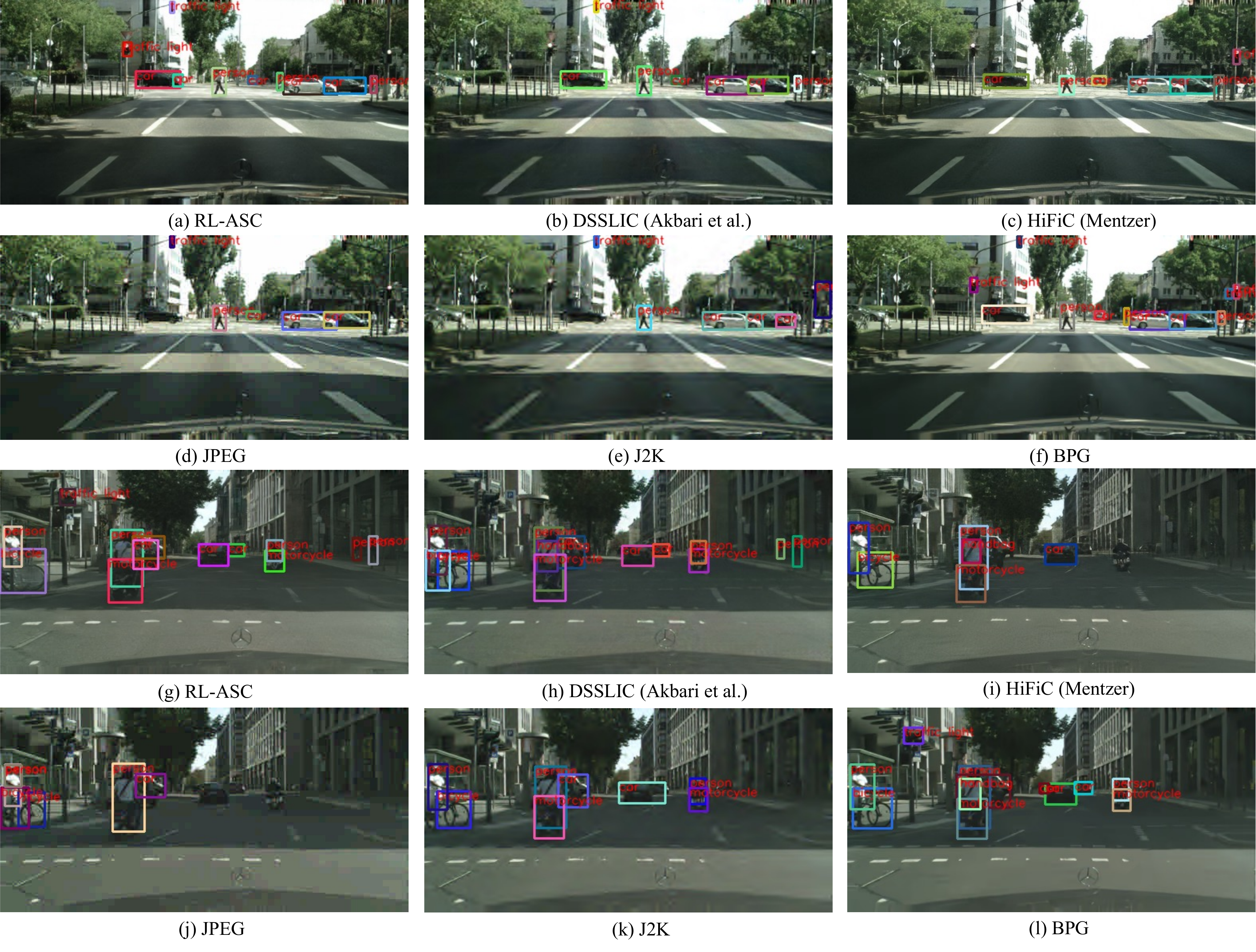}
\caption{ Examples of image coding on downstream object detection task. Two randomly selected images are given as examples. The bounding boxes and labels of the decompressed images of different image codecs are produced by the pretrained MaskRCNN. }
\label{detection}
\end{figure*}

The input images, ground truths and the semantic label maps predicted on the decompressed images are shown in Fig. \ref{segmentation}. We can observe that the semantic concepts of the proposed RL-ASC can be well recognized and localized on the decompressed images (a) and (g). The segmentation results are similar to the ground truths, which validates that the proposed RL-ASC could  maintain the overall meanings of the input image and suffer from little semantic information loss. The deep learning-based codecs DSSLIC \cite{Akbari} and HiFiC \cite{mentzer2020high} perform better than the classic engineered codecs \cite{BPG, JPEG2000, JPEG}. The semantic labels (b), (c), (h), and (i) represent major concepts such as car, building, and road, while the less important information is missing. As shown in (d), (e), (f), (j), (k), and (i), the outputs of JPEG \cite{JPEG}, J2K \cite{JPEG2000}, and BPG \cite{BPG} fail to conduct semantic segmentation task, and the semantic concepts are misinterpreted by the downstream task. It is because the decompressed image of the engineered codecs degrade heavily at low bitrate, suffering from blocky, blurring or ringing artifact.

The object detection performance on the decompressed images of the proposed RL-ASC as well as the baselines are illustrated in Fig. \ref{detection}. It is observed that the decompressed images (a) and (g) of the proposed RL-ASC preserve the objects comprehensively. Small objects such as traffic lights and overlapped cars can be detected properly on (a) and (g), which validates the incredible semantic exchange ability of the proposed RL-ASC method. However, since the baselines attempt to recover the exact pixels while ignoring the global underlying meanings, some reconstructed objects fail to be detected accurately. For instance, the traffic light is not detected on HiFiC  decompressed image (c) and DSSLIC decompressed image (h), and larger objects such as cars and persons may even be miss detected on classic engineered codecs \cite{BPG, JPEG2000, JPEG}.

\begin{figure*}[t]
\centering
\includegraphics[width=1.0\linewidth]{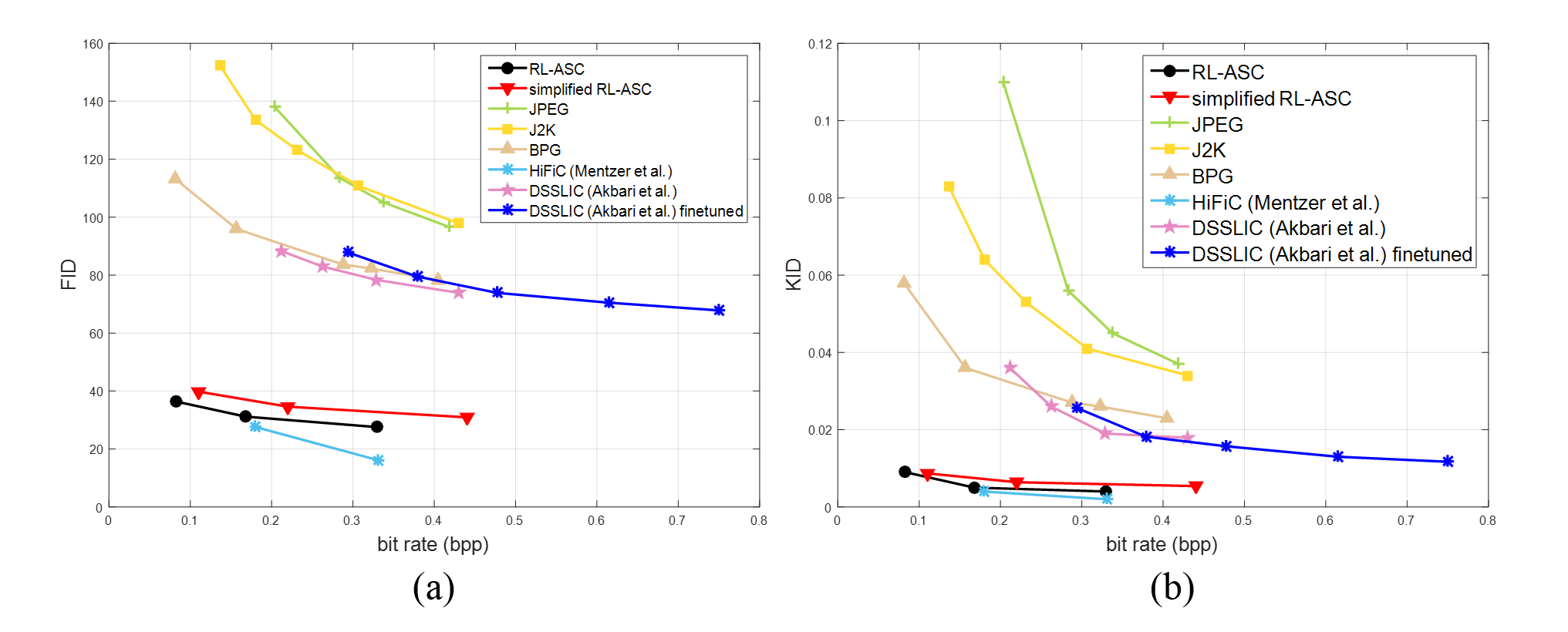}
\caption{The perceptual performance of different image codecs in terms of (a) FID and (b) KID. The lower value means better performance.}
\label{perceptual}  
\end{figure*}

\begin{figure*}[t]
\centering
\includegraphics[width=0.8\linewidth]{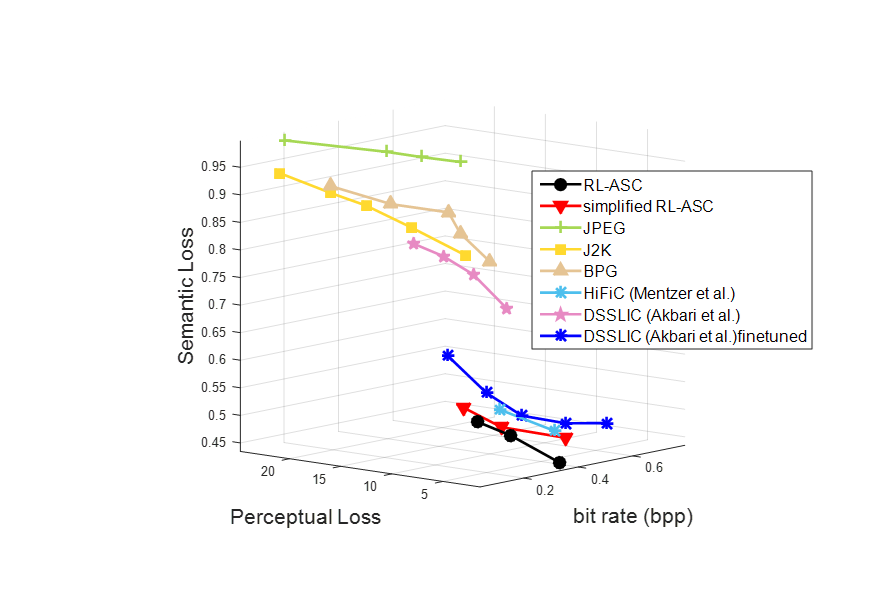}
\caption{The rate-semantic-perceptual curves of different image codecs. At a certain bitrate, lower semantic loss and lower perceptual loss means better performance.}
\label{TTO}
\end{figure*}

\begin{figure*}[t]
\centering
\includegraphics[width=0.9\linewidth]{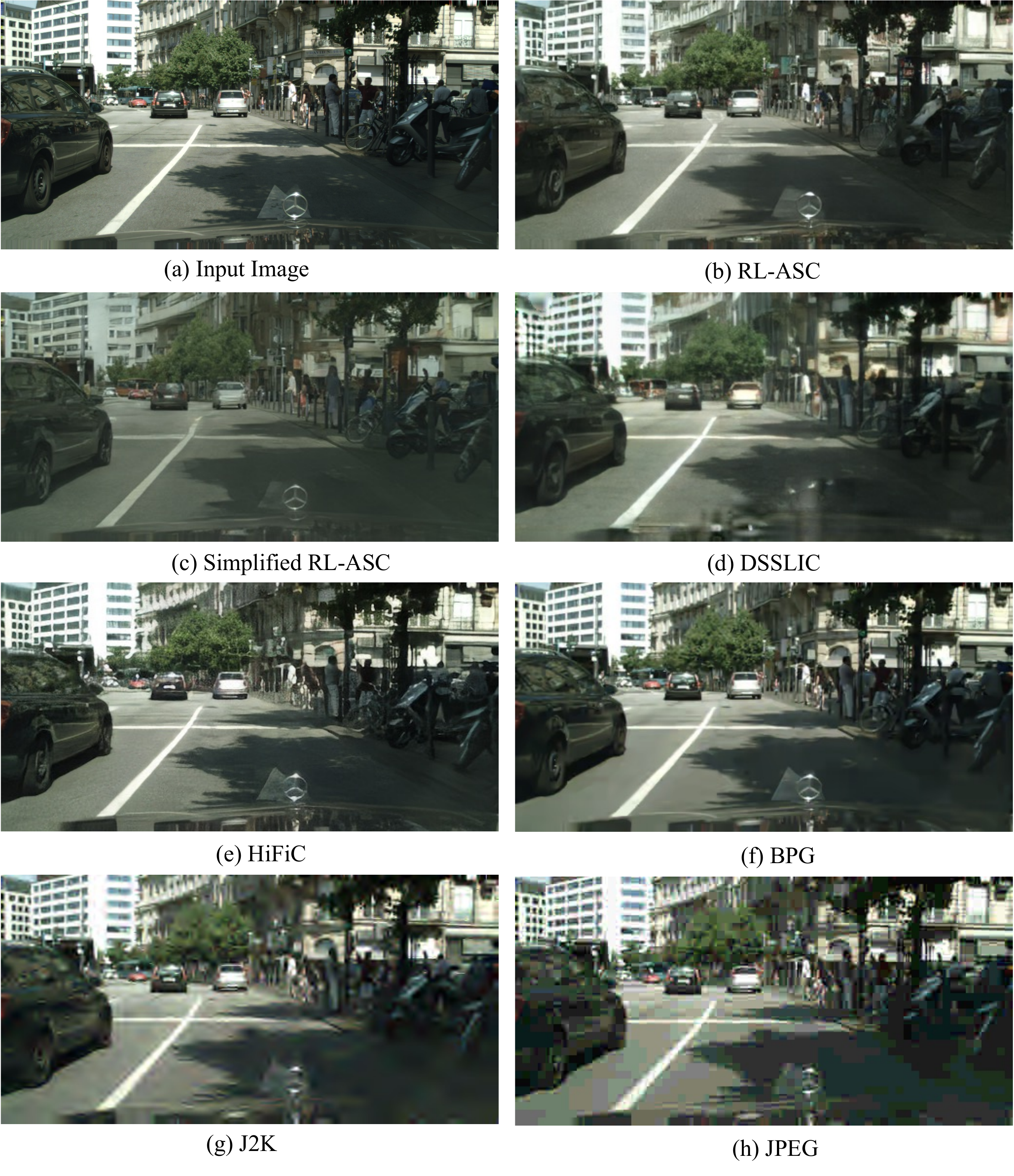}
\caption{A randomly selected input image and the reconstructed images of different image codecs at similar bitrate (0.16 bpp).}
\label{reconst}
\end{figure*}

\begin{figure*}[t]
\centering
\includegraphics[width=1.0\linewidth]{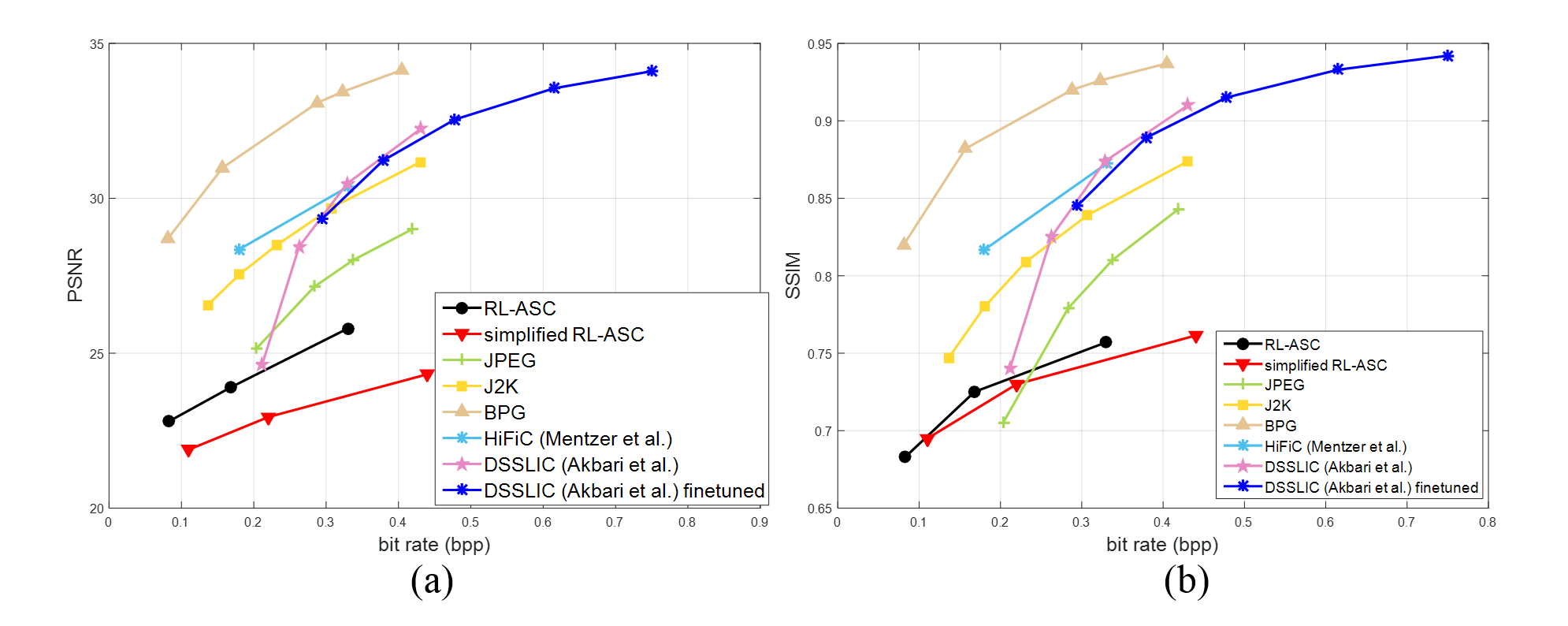}
\caption{The rate-distortion performance in terms of PSNR and SSIM of different image codecs. Higher value means better performance.}
\label{classic}
\end{figure*}

\subsection{Perceptual Performance}
We compare the perceptual performance of the proposed RL-ASC and simplified RL-ASC with the baselines at different bitrates.

\subsubsection{Objective Quality} 
The perceptual performance of different image codecs in terms of FID \cite{Heusel} and KID  \cite{Binkowski} under different bitrates is illustrated in Fig. \ref{perceptual}. The proposed RL-ASC is comparable with HiFiC \cite{mentzer2020high} and outperforms other baselines by a large margin. This can be interpreted as the proposed method incorporates GANs architecture and adopts adversarial loss that enforces the reconstructed image to be natural and realistic, which also validates the effectiveness of the well-designed semantic encoder and semantic decoder. In addition, the RL-based semantic bit allocation model results in convincing increase on perceptual performance under the same bitrate compared to the simplified RL-ASC. There is a moderate degradation for the finetuned DSSLIC compared to the original DSSLIC model in terms of FID and KID. This can be interpreted that the finetuned model emphasizes more on semantic fidelity and sacrifice perceptual performance.

We can further illustrate the triple trade-off rate-semantic-perceptual of the proposed RL-ASC method and the baselines in Fig. \ref{TTO}. In particular, the semantic loss is defined as (1-mIoU), considering the semantic segmentation task. The perceptual loss is defined as the weighted addition of FID and KID values, where a lower value means better performance. As shown in Fig. \ref{TTO}, the proposed RL-ASC achieves lower semantic and perceptual loss compared to the baselines at equal bitrate by a large margin, which validates the effectiveness of the proposed method in semantic exchange and visual performance. Note that the deep learning based methods RL-ASC, \cite{mentzer2020high, Akbari} outperform the classic engineered codecs, which can be interpreted that the deep features account for better image understanding. Moreover, the DSSLIC finetuned by the semantic loss achieves great progress in this triplet loss, compared to the original DSSLIC model.

\subsubsection{Subjective Quality}
The reconstructed images of different image codecs, as well as the original randomly selected input image, are shown in Fig. \ref{reconst}. To ensure a fair comparison, the RL-ASC encodes the image at a bitrate 0.16 bpp while the baselines encode the image at an equal or higher bitrate. It can be observed that the decompressed image (b) of the proposed RL-ASC is almost indistinguishable from the input image (a) even at such a low bitrate. Compared to the simplified RL-ASC (c), the bits in (b) concentrate on salient semantic concepts and therefore result in visually pleasant reconstruction. In addition, the baselines suffer from blur, ringing, and blocky artifacts at low bitrate, as shown in (d), (f), (g), and (h).

\begin{figure*}[t]
\centering
\includegraphics[width=0.5\linewidth]{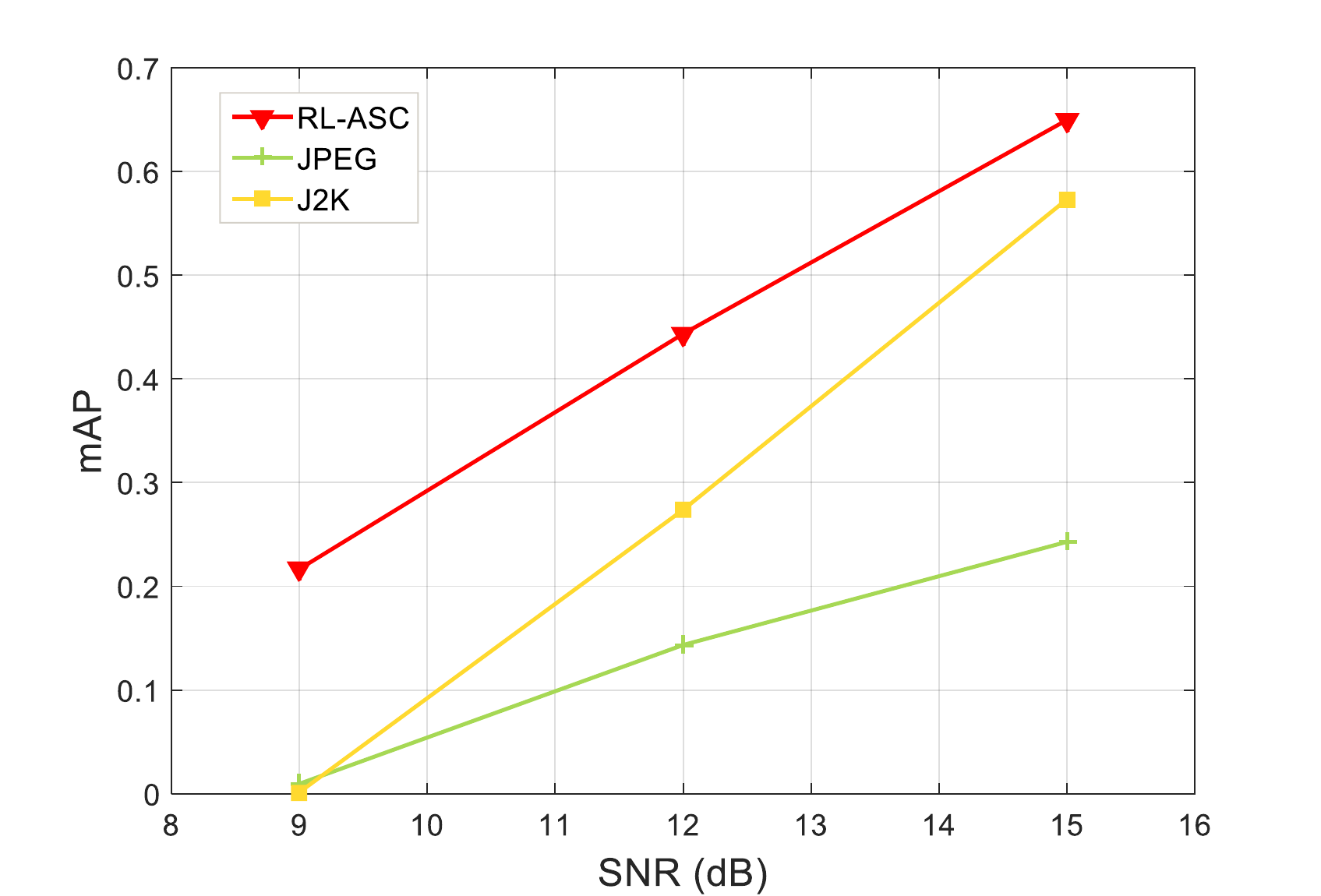}
\caption{mAP score versus SNR for the same bitrate of compressed images (0.33 bpp) under the AWGN channel.}
\label{SNR}
\end{figure*}

\subsection{Rate-Distortion Performance}
We evaluate the rate-distortion performance of the proposed RL-ASC, simplified RL-ASC, and the baselines in terms of PSNR and SSIM, which are the most widely used metric in the traditional image coding system that attempts to minimize symbol error. As shown in Fig. \ref{classic}, the baselines achieve better performance since they are optimized with the PSNR or SSIM metric. On the contrary, the proposed method is tolerable to pixel errors and does not attempt to ensure local consistency. It can be observed that the RL-ASC outperforms the simplified RL-ASC since the former represents complex objects with higher precision and the rest of the image is simple to be encoded. Moreover, a moderate degradation is occurred in DSSLIC-finetuned compared to DSSLIC in terms of PSNR and SSIM, which validates the trade-off between the classic pixel level loss and the novel semantic loss.

\subsection{Anti-Noise Performance}
In this last example, we consider the Additive White Gaussian Noise (AWGN) channel and evaluate the robustness of the proposed RL-ASC as well as the classical codecs \cite{JPEG, JPEG2000} to physical noise. Fig. \ref{SNR} shows the object detection performance on the decompressed images in terms of mAP score in different SNRs. Note that the same compression rate (0.33 bpp) is adopted for different methods. The proposed RL-ASC outperforms the baselines by a large margin, which demonstrates the robustness of the proposed method in ISC scenario. The baselines lost most of the information in $SNR=9$, and the reconstructed image is uninterpretable with mAP $\approx$ 0. For $SNR=15$ or higher, the effect of the physical noise is ignorable, since the mAP accuracy approximates that of the ideal channel condition. In this case, the object detection result obtained by the pretrained MaskRCNN on the input image is deemed as the ground truth label. A more similar detection result on the reconstructed image leads to higher mAP score.

\section{Conclusion}
In this paper, we considered the ISC system and presented a deep learning-based semantic image coding approach that interprets and encodes images beyond pixel level. We first proposed the novel rate-semantic-perceptual criterion to integrate the semantic fidelity and perceptual quality in the optimization process of the semantic coding. Accordingly, we designed the semantic concept as the novel representation unit and proposed a convolutional semantic encoder to extract semantic information. Driven by the semantic analysis task such as object detection or semantic segmentation, an RL-based semantic bit allocation model is presented to realize the optimization criterion and encode each semantic concept with adaptive quantization.  At the receiver side, a generative semantic decoder that adopts attention model to fuse the local and global features is designed to reconstruct the semantic concepts. With the extracted semantic information, the proposed RL-ASC can facilitate multiple vision tasks in the semantic communication scenario.  We compared the decompressed samples of the proposed approach with that of the baselines and showed that FID, KID, and mIoU can be valuable tools to better predict human preferences and the efficiency of semantic exchange. The experiments demonstrated the ability of RL-ASC to produce reconstructions with high semantic similarity, naturalness, and remarkably reduced transmission data amount. Also, the proposed RL-ASC is robust to noise in AWGN channel.

%


%






\ifCLASSOPTIONcaptionsoff
  \newpage
\fi



%
\linespread{1.2}
\bibliographystyle{IEEEtran}
\bibliography{reference}

\begin{thebibliography}{10}

\bibitem{BPG}
\url{https://bellard.org/bpg/.}

\bibitem{Agustsson}
Eirikur Agustsson, Michael Tschannen, Fabian Mentzer, Radu Timofte, and Luc~Van
  Gool.
\newblock Generative adversarial networks for extreme learned image
  compression.
\newblock In {\em Int. Conf. Comput. Vis. (ICCV)}, pages 221--231, 2019.

\bibitem{Akbari}
Mohammad Akbari, Jie Liang, and Jingning Han.
\newblock Dsslic: deep semantic segmentation-based layered image compression.
\newblock In {\em 2019-2019 IEEE Int. Conf. Acoust., Speech, Signal Process.
  (ICASSP)}, pages 2042--2046. IEEE, 2019.

\bibitem{ICLR}
Johannes Ball{\'e}, Valero Laparra, and {Eero P.} Simoncelli.
\newblock End-to-end optimized image compression.
\newblock In {\em 5th Int. Conf. on Learning Representations, ICLR 2017}, 2017.

\bibitem{Binkowski}
Miko{\l}aj Bi{\'n}kowski, Dougal~J Sutherland, Michael Arbel, and Arthur
  Gretton.
\newblock Demystifying {MMD GANs}.
\newblock {\em arXiv preprint arXiv:1801.01401}, 2018.

\bibitem{Bjntegaard2001CalculationOA}
Gisle Bj{\o}ntegaard.
\newblock Calculation of average {PSNR} differences between {RD}-curves.
\newblock In {\em ITU-T VCEG-M33, April,2001}, 2001.

\bibitem{Blau}
Yochai Blau and Tomer Michaeli.
\newblock The perception-distortion tradeoff.
\newblock In {\em IEEE Conf. Comput. Vis. Pattern Recog. (CVPR)}, pages
  6228--6237, 2018.

\bibitem{Bourtsoulatze}
Eirina Bourtsoulatze, David~Burth Kurka, and Deniz G{\"u}nd{\"u}z.
\newblock Deep joint source-channel coding for wireless image transmission.
\newblock {\em IEEE Trans. on Cogn. Commun. Netw.}, 5(3):567--579, 2019.

\bibitem{chen2019learning}
Zhibo Chen and Tianyu He.
\newblock Learning based facial image compression with semantic fidelity
  metric.
\newblock {\em Neurocomputing}, 338:16--25, 2019.

\bibitem{20204409437480}
Zhengxue Cheng, Heming Sun, Masaru Takeuchi, and Jiro Katto.
\newblock Learned image compression with discretized gaussian mixture
  likelihoods and attention modules.
\newblock In {\em IEEE Conference Conf. Comput. Vis. and Pattern Recog.,
  (CVPR)}, pages 7936 -- 7945, 2020.

\bibitem{Cisco2017Cisco}
Cisco.
\newblock Cisco visual networking index: Global mobile data traffic forecast
  update 2017-2022.
\newblock 2017.

\bibitem{Cordts}
Marius Cordts, Mohamed Omran, Sebastian Ramos, Timo Rehfeld, Markus Enzweiler,
  Rodrigo Benenson, Uwe Franke, Stefan Roth, and Bernt Schiele.
\newblock The cityscapes dataset for semantic urban scene understanding.
\newblock In {\em IEEE Conf. Comput. Vis. Pattern Recog. (CVPR)}, pages
  3213--3223, 2016.

\bibitem{7498955}
Samuel Dodge and Lina Karam.
\newblock Understanding how image quality affects deep neural networks.
\newblock In {\em Int. Conf. Qual. Multimed. Exp. (QoMEX)}, pages 1--6, 2016.

\bibitem{8461983}
Nariman Farsad, Milind Rao, and Andrea Goldsmith.
\newblock Deep learning for joint source-channel coding of text.
\newblock In {\em 2018 IEEE Int. Conf. on Acoustics, Speech and Signal
  Processing (ICASSP)}, pages 2326--2330, 2018.

\bibitem{2019Differentiable}
R.~Gong, X.~Liu, S.~Jiang, T.~Li, and J.~Yan.
\newblock Differentiable soft quantization: Bridging full-precision and low-bit
  neural networks.
\newblock In {\em IEEE Int. Conf. on Comput. Vis. (ICCV)}, 2019.

\bibitem{goodfellow2014generative}
Ian Goodfellow, Jean Pouget-Abadie, Mehdi Mirza, Bing Xu, David Warde-Farley,
  Sherjil Ozair, Aaron Courville, and Yoshua Bengio.
\newblock Generative adversarial nets.
\newblock {\em Adv. Neural Inform. Process. Syst.}, 27, 2014.

\bibitem{Grm}
Klemen Grm, Vitomir Štruc, Anais Artiges, Matthieu Caron, and Hazım~K.
  Ekenel.
\newblock Strengths and weaknesses of deep learning models for face recognition
  against image degradations.
\newblock {\em IET Biom.}, 7(1):81--89, 2018.

\bibitem{he2018mask}
Kaiming He, Georgia Gkioxari, Piotr Dollar, and Ross Girshick.
\newblock Mask {R-CNN}.
\newblock In {\em Proceedings of the IEEE Int. Conf. on Comput. Vis. (ICCV)},
  Oct 2017.

\bibitem{2019Beyond}
T.~He, S.~Sun, Z.~Guo, and Z.~Chen.
\newblock Beyond coding: Detection-driven image compression with semantically
  structured bit-stream.
\newblock In {\em 2019 Picture Coding Symposium (PCS)}, 2019.

\bibitem{Heusel}
Martin Heusel, Hubert Ramsauer, Thomas Unterthiner, Bernhard Nessler, and Sepp
  Hochreiter.
\newblock {GANs} trained by a two time-scale update rule converge to a local
  nash equilibrium.
\newblock {\em Neural Info. Process. Systems (NIPS)}, page 6629–6640, 2017.

\bibitem{Hu2020CoarsetoFineHM}
Yueyu Hu, Wenhan Yang, and Jiaying Liu.
\newblock Coarse-to-fine hyper-prior modeling for learned image compression.
\newblock In {\em AAAI}, 2020.

\bibitem{johnson2016perceptual}
Justin Johnson, Alexandre Alahi, and Li~Fei-Fei.
\newblock Perceptual losses for real-time style transfer and super-resolution.
\newblock In {\em Eur. Conf. Comput. Vis.}, pages 694--711. Springer, 2016.

\bibitem{20191106642196}
Nick Johnston, Damien Vincent, David Minnen, Michele Covell, Saurabh Singh,
  Troy Chinen, Sung Jin~Hwang, Joel Shor, and George Toderici.
\newblock Improved lossy image compression with priming and spatially adaptive
  bit rates for recurrent networks.
\newblock In {\em IEEE Conf. Comput. Vis. and Pattern Recog., (CVPR)}, pages
  4385 -- 4393, 2018.

\bibitem{kingma2014adam}
Diederik~P Kingma and Jimmy Ba.
\newblock Adam: A method for stochastic optimization.
\newblock {\em Computer Science}, 2014.

\bibitem{Kurka}
David~Burth Kurka and Deniz G{\"u}nd{\"u}z.
\newblock Bandwidth-agile image transmission with deep joint source-channel
  coding.
\newblock {\em {IEEE Trans. on Wireless Commun.},}, 2020.

\bibitem{20193507365934}
Jooyoung Lee, Seunghyun Cho, and Seung-Kwon Beack.
\newblock Context-adaptive entropy model for end-to-end optimized image
  compression.
\newblock In {\em 7th International Conference on Learning Representations,
  ICLR}, 2019.

\bibitem{LI202213}
Binglin Li, Linwei Ye, Jie Liang, Yang Wang, and Jingning Han.
\newblock Region-of-interest and channel attention-based joint optimization of
  image compression and computer vision.
\newblock {\em Neurocomputing}, 500:13--25, 2022.

\bibitem{9472999}
Xin Li, Jun Shi, and Zhibo Chen.
\newblock Task-driven semantic coding via reinforcement learning.
\newblock {\em IEEE Trans. Image Process.}, 30:6307--6320, 2021.

\bibitem{liu2019classification}
Dong Liu, Haochen Zhang, and Zhiwei Xiong.
\newblock On the classification-distortion-perception tradeoff.
\newblock {\em Advances in Neural Information Processing Systems}, 2019.

\bibitem{lu2021reinforcement}
Kun Lu, Rongpeng Li, Xianfu Chen, Zhifeng Zhao, and Honggang Zhang.
\newblock Reinforcement learning-powered semantic communication via semantic
  similarity.
\newblock {\em arXiv preprint arXiv:2108.12121}, 2021.

\bibitem{2018Conditional}
F.~Mentzer, E.~Agustsson, M.~Tschannen, R.~Timofte, and L~Van Gool.
\newblock Conditional probability models for deep image compression.
\newblock In {\em IEEE Conf. Comput. Vis. and Pattern Recog. (CVPR)}, 2018.

\bibitem{mentzer2020high}
Fabian Mentzer, George Toderici, Michael Tschannen, and Eirikur Agustsson.
\newblock High-fidelity generative image compression.
\newblock {\em Advances in Neural Information Processing Systems}, 2020.

\bibitem{2018Nips}
David Minnen, Johannes Balle, and George. Toderici.
\newblock Joint autoregressive and hierarchical priors for learned image
  compression.
\newblock In {\em Advances in Neural Information Processing Systems.},
  volume~31, 2018.

\bibitem{Park}
Taesung Park, Ming-Yu Liu, Ting-Chun Wang, and Jun-Yan Zhu.
\newblock Semantic image synthesis with spatially-adaptive normalization.
\newblock In {\em IEEE Conf. Comput. Vis. Pattern Recog. (CVPR)}, pages
  2337--2346, 2019.

\bibitem{DBLP:journals/corr/abs-2201-01389}
Zhijin Qin, Xiaoming Tao, Jianhua Lu, and Geoffrey~Ye Li.
\newblock Semantic communications: Principles and challenges.
\newblock {\em CoRR}, abs/2201.01389, 2022.

\bibitem{JPEG2000}
Majid Rabbani.
\newblock {JPEG2000}: Image compression fundamentals, standards and practice.
\newblock {\em J Electron Imaging}, 11(2):286, 2002.

\bibitem{pmlr-v70-rippel17a}
Oren Rippel and Lubomir Bourdev.
\newblock Real-time adaptive image compression.
\newblock In Doina Precup and Yee~Whye Teh, editors, {\em 34th Int. Conf. Mach.
  Learn. (ICML)}, volume~70 of {\em Proceedings of Machine Learning Research},
  pages 2922--2930. PMLR, 06--11 Aug 2017.

\bibitem{2018MobileNetV2}
M.~Sandler, A.~Howard, M.~Zhu, A.~Zhmoginov, and L.~C. Chen.
\newblock Mobilenetv2: Inverted residuals and linear bottlenecks.
\newblock In {\em 2018 IEEE/CVF Conference on Computer Vision and Pattern
  Recognition (CVPR)}, 2018.

\bibitem{Santurkar}
Shibani Santurkar, David Budden, and Nir Shavit.
\newblock Generative compression.
\newblock In {\em 2018 Picture Coding Symp. (PCS)}, pages 258--262. IEEE, 2018.

\bibitem{1949The}
Claude~E. Shannon and W.~Weaver.
\newblock The mathematical theory of communication.
\newblock {\em Philosophical Review}, 60(3), 1949.

\bibitem{2021Variable}
M.~Song, J.~Choi, and B.~Han.
\newblock Variable-rate deep image compression through spatially-adaptive
  feature transform.
\newblock In {\em Int. Conf. on Comput. Vis.}, 2021.

\bibitem{9281078}
Simeng Sun, Tianyu He, and Zhibo Chen.
\newblock Semantic structured image coding framework for multiple intelligent
  applications.
\newblock {\em IEEE Trans. Circuits Syst. Video Technol.}, 31(9):3631--3642,
  2021.

\bibitem{Theis}
Lucas Theis, Wenzhe Shi, Andrew Cunningham, and Ferenc Husz{\'a}r.
\newblock Lossy image compression with compressive autoencoders.
\newblock {\em arXiv preprint arXiv:1703.00395}, 2017.

\bibitem{2015Variable}
G.~Toderici, S.~M. O'Malley, S.~J. Hwang, D.~Vincent, D.~Minnen, S.~Baluja,
  M.~Covell, and R.~Sukthankar.
\newblock Variable rate image compression with recurrent neural networks.
\newblock {\em Computer Science}, 2015.

\bibitem{20181304961211}
George Toderici, Damien Vincent, Nick Johnston, Sung~Jin Hwang, David Minnen,
  Joel Shor, and Michele Covell.
\newblock Full resolution image compression with recurrent neural networks.
\newblock In {\em IEEE Conference Conf. Comput. Vis. and Pattern Recog.,
  (CVPR)}, volume 2017-January, pages 5435 -- 5443, 2017.

\bibitem{torfason2018towards}
Robert Torfason, Fabian Mentzer, Eirikur Agustsson, Michael Tschannen, Radu
  Timofte, and Luc Van~Gool.
\newblock Towards image understanding from deep compression without decoding.
\newblock {\em 6th Int. Conf. on Learning Representations, (ICLR)}, 2018.

\bibitem{JPEG}
Gregory~K Wallace.
\newblock The {JPEG} still picture compression standard.
\newblock {\em IEEE Trans. Consum. Electron.}, 38(1):xviii--xxxiv, 1992.

\bibitem{Weng}
Zhenzi Weng, Zhijin Qin, and Geoffrey~Ye Li.
\newblock Semantic communications for speech signals.
\newblock {\em IEEE Int. Conf. on Commun. (ICC)}, 2020.

\bibitem{Xie}
Huiqiang Xie and Zhijin Qin.
\newblock A lite distributed semantic communication system for internet of
  things.
\newblock {\em IEEE J. Sel. Areas Commun.}, 39(1):142--153, 2020.

\bibitem{XieTSP}
Huiqiang Xie, Zhijin Qin, Geoffrey~Ye Li, and Biing-Hwang Juang.
\newblock Deep learning enabled semantic communication systems.
\newblock {\em IEEE Trans. Signal Process.}, 69:2663--2675, 2021.

\bibitem{XuTCSVT}
Jialong Xu, Bo~Ai, Wei Chen, Ang Yang, Peng Sun, and Miguel Rodrigues.
\newblock Wireless image transmission using deep source channel coding with
  attention modules.
\newblock {\em IEEE Trans. Circuits Syst. Video Technol.}, 2021.

\bibitem{Yu_2016}
Jiahui Yu, Yuning Jiang, Zhangyang Wang, Zhimin Cao, and Thomas Huang.
\newblock Unitbox: An advanced object detection network.
\newblock In {\em Proceedings of the 24th {ACM} int. conf. on Multimedia}.
  {ACM}, 2016.

\bibitem{Zhao}
Hengshuang Zhao, Jianping Shi, Xiaojuan Qi, Xiaogang Wang, and Jiaya Jia.
\newblock Pyramid scene parsing network.
\newblock In {\em IEEE Conf. Comput. Vis. Pattern Recog. (CVPR)}, pages
  2881--2890, 2017.

\end{thebibliography}

%








\end{document}